\documentclass{article}
\usepackage{ijcai25}

\usepackage{times}
\usepackage{soul}
\usepackage{url}
\usepackage[hidelinks]{hyperref}
\usepackage[utf8]{inputenc}
\usepackage[small]{caption}
\usepackage{graphicx}
\usepackage{multirow}
\usepackage{array}
\usepackage{amsmath}
\usepackage{mathtools}
\usepackage{amssymb}
\usepackage{amsthm}
\usepackage{booktabs}
\usepackage[table,xcdraw]{xcolor}
\usepackage{algorithm}
\usepackage{algorithmic}
\usepackage[switch]{lineno}

\title{MC3D-AD: A Unified Geometry-aware Reconstruction Model for Multi-category 3D Anomaly Detection}

\author{
Jiayi Cheng$^1$
\and
Can Gao$^{1,4}$\footnote{Corresponding author}\and
Jie Zhou$^{2,3,4}$\and
Jiajun Wen$^1$\and
Tao Dai$^1$\And
Jinbao Wang$^{2,3,4}$
\\
\affiliations
$^1$College of Computer Science and Software Engineering, Shenzhen University\\
$^2$School of Artificial Intelligence, Shenzhen University\\
$^3$National Engineering Laboratory for Big Data System Computing Technology, Shenzhen University\\
$^4$Guangdong Provincial Key Laboratory of Intelligent Information Processing, Shenzhen University\\
\emails
2005gaocan@163.com
}

\begin{document}

\maketitle

\begin{abstract}
3D Anomaly Detection (AD) is a promising means of controlling the quality of manufactured products. However, existing methods typically require carefully training a task-specific model for each category independently, leading to high cost, low efficiency, and weak generalization.
This study presents a novel unified model for Multi-Category 3D Anomaly Detection (MC3D-AD) that aims to utilize both local and global geometry-aware information to reconstruct normal representations of all categories. First, to learn robust and generalized features of different categories, we propose an adaptive geometry-aware masked attention module that extracts geometry variation information to guide mask attention. Then, we introduce a local geometry-aware encoder reinforced by the improved mask attention to encode group-level feature tokens. Finally, we design a global query decoder that utilizes point cloud position embeddings to improve the decoding process and reconstruction ability. This leads to local and global geometry-aware reconstructed feature tokens for the 3D AD task. MC3D-AD is evaluated on two publicly available Real3D-AD and Anomaly-ShapeNet datasets, and exhibits significant superiority over current state-of-the-art single-category methods, achieving 3.1\% and 9.3\% improvement in object-level AUROC over Real3D-AD and Anomaly-ShapeNet, respectively. The code is available at \url{https://github.com/iCAN-SZU/MC3D-AD}.
\end{abstract}

\section{Introduction}
Anomaly Detection (AD) is a critical task for quality control in the manufacturing industry. Early research has concentrated on 2D image data and has achieved significant advancements~\cite{DRAEM,DeSTSeg2023,uniad,Lu2023hvq}. With the increasing demand for high-precision industrial products, 3D-AD~\cite{Bergmann_2022,real3dad} has garnered growing attention from researchers, and its objective is to identify and localize anomalous points or regions from 3D point cloud data. 

Point cloud data exhibits the characteristics of disorder, sparseness, and structurelessness, which pose great challenges for anomaly detection. Existing 3D-AD methods~\cite{PO3AD} can generally be categorized into: feature embedding-based and reconstruction-based ones. For feature embedding-based approaches, methods like Reg3D-AD~\cite{real3dad} and Group3-AD~\cite{zhu2024towards} have demonstrated their effectiveness by extracting feature embeddings from normal samples. While reconstruction-based methods, such as IMRNet~\cite{IMRNet} and R3DAD~\cite{Zhou2024R3D}, focus on learning key features to restore point cloud data, achieving anomaly detection by calculating reconstruction errors. 

Additionally, in light of the rich information hidden in multimodal data, multimodal 3D-AD has also attracted much attention from researchers. Some methods, such as BTF~\cite{Horwitz2023BTF}, extract statistical information from both RGB and depth modules to perform anomaly detection. Recently, deep learning-based approaches, such as M3DM~\cite{Wang2023multimodal} and CMPF~\cite{cao2023complementarypseudomultimodalfeature}, have shown promising results by learning feature representations from multimodal data. 

\begin{figure}[t!]
    \centering
    \includegraphics[width=\linewidth]{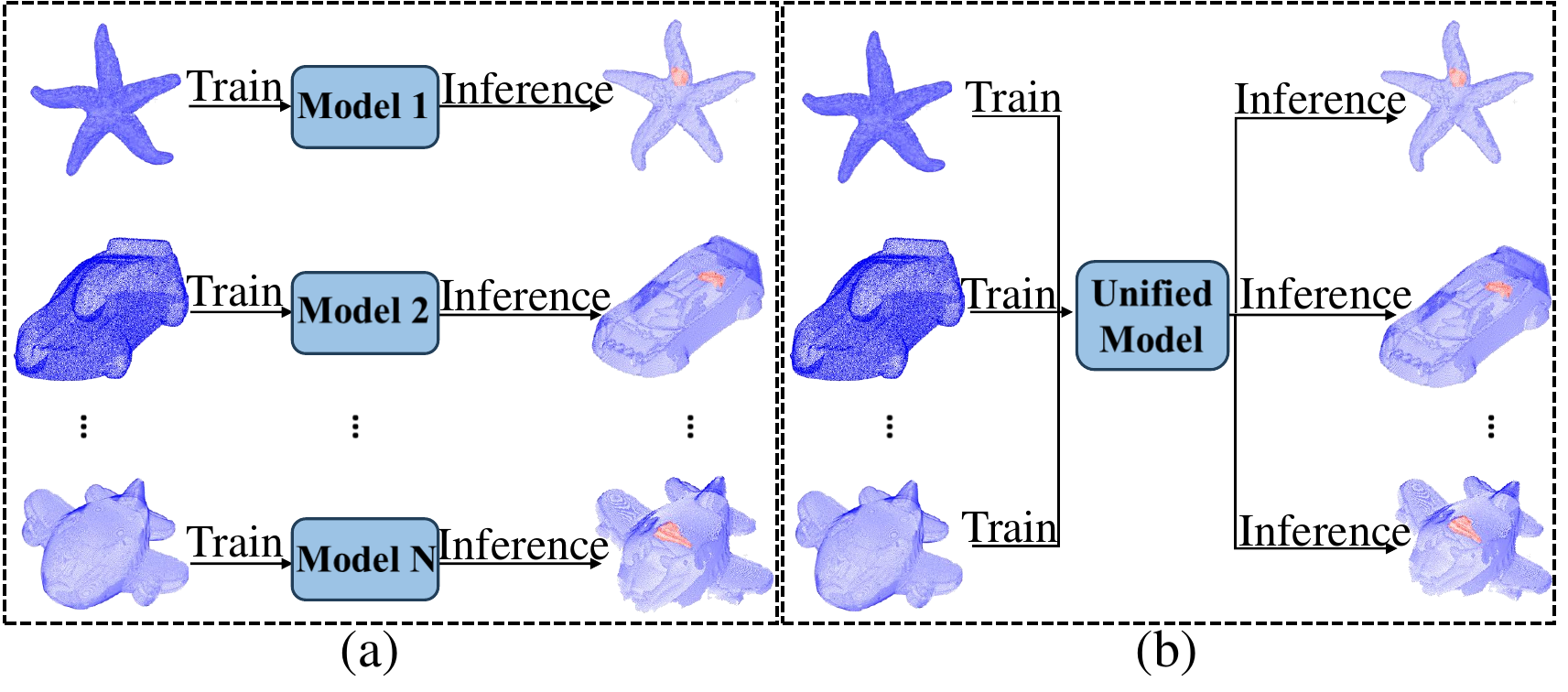}
    \caption{Different settings for 3D anomaly detection. (a) Single-category anomaly detection; (b) Multi-category anomaly detection.}
    \label{fig1}
\end{figure}

Despite achieving very appealing results, some challenges still exist: (1) The trained models are task-specific and lack the generalization to different tasks. In other words, they are required to train an individual and distinct model for each category, seriously limiting their practicality. (2) Existing reconstruction-based methods may fail to learn the intrinsic features for reconstruction, leading to the problem of ``identity shortcut", where the input is directly copied for output without considering its content~\cite{uniad}. As verified in 2D images~\cite{uniad,Diad,Lu2023hvq}, this phenomenon is significantly amplified in the setting of multi-category anomaly detection. To address these challenges, we propose a unified geometry-aware reconstruction model for 3D anomaly detection named MC3D-AD. Different from previous methods which require training a privately owned model for each category, our method aims to train only one unified model to perform 3D anomaly detection for all categories (See Figure~\ref{fig1}). 

In light of that point cloud anomalies are usually manifested as irregularities and abnormalities in local geometry structures, we introduce an adaptive geometry-aware masked attention to improve the local feature representation for different categories. It explicitly computes geometric variations within the neighborhood of points and intentionally masks key features selected using their quantified geometric information. To accurately identify anomalies from point cloud data, a transformer-based~\cite{transformer} architecture is adopted, which incorporates point cloud position embeddings~\cite{Li_2023_CVPR} as global queries and the adaptive geometry-aware masked attention to reconstruct feature tokens for multi-category anomaly detection and localization.

The main contributions of this paper are summarized as follows:
\begin{itemize}
    \item To perform multi-category 3D anomaly detection, we present a unified framework based on feature reconstruction. To the best of our knowledge, it is the first time to explore multi-category 3D anomaly detection by training only one model.
   
    \item To learn robust and generalized representations across categories, we propose a novel adaptive geometry-aware masked attention, which explicitly captures neighborhood geometry information for representation, facilitating the extraction of reconstruction features and also enhancing the interpretability of the model.
    
    \item To achieve accurate anomaly detection, we design a local and global geometry-aware transformer, which is reinforced by the proposed adaptive geometry-aware masked attention, thereby providing the ability to reconstruct point clouds from different categories.
    
    \item Extensive experiments are conducted to verify the effectiveness of the proposed model, and very impressive results are achieved, with an object-level AUROC improvement of 3.1\% over the state-of-the-arts single-category model on Real3D-AD and 9.3\% on Anomaly-ShapeNet, respectively.
\end{itemize}

\section{Related Work}
\subsection{Feature Embedding-Based Methods}
Feature embedding-based methods\cite{LookLiang} extract features from normal samples to form a memory bank using pre-trained models and identify anomalies by comparing test sample features with those in the memory bank. Reg3D-AD~\cite{real3dad} used the pre-trained PointMAE~\cite{Pang2022pointmae} to extract normal features from registered point cloud data and constructed a memory bank to store both global geometric and local coordinate features for anomaly detection. Group3AD~\cite{zhu2024towards} introduced contrastive learning for clustering groups to ensure intra-cluster compactness and inter-cluster uniformity, leveraging group-level features stored in a memory bank to detect anomalies. 3D-ST~\cite{Bergmann_2023_3DST} adopted a student-teacher framework to perform feature matching between two networks for 3D anomaly detection. M3DM~\cite{Wang2023multimodal} utilized contrastive learning to align RGB and depth modalities, creating a fused representation and a three-level memory bank to jointly enhance detection performance. CPMF~\cite{cao2023complementarypseudomultimodalfeature} projected 3D point clouds into 2D images from multi-view and considered image features as global semantic information to complement 3D features, thereby establishing a pseudo multimodal memory bank for anomaly detection. PointAD~\cite{pointad2024} aligned local and global features extracted from 2D projections of 3D point clouds using a pre-trained vision-language model, enabling zero-shot 3D anomaly detection.

\subsection{Reconstruction-Based Methods}
Reconstruction-based methods try to encode normal point cloud data into informative feature representation and restore these features into the original form, with points exhibiting high reconstruction errors identified as anomalies. IMRNet~\cite{IMRNet} enhanced PointMAE by incorporating geometric-preserving downsampling and random masking to improve reconstruction fidelity for anomaly detection. R3DAD~\cite{Zhou2024R3D} leveraged PointNet to iteratively reconstruct fully masked point clouds using a diffusion process, enabling precise localization of abnormal regions. Shape-Guided~\cite{shape2023chu} introduced dual memory banks to store normal features extracted from RGB and 3D modalities and reconstructed the input sample at the feature level to achieve robust anomaly detection. Although achieving encouraging results, these methods need to train a task-specific model for each category. Therefore, it is highly desired to develop a unified all-in-one model for all categories.

\section{The Proposed Approach}
\subsection{Problem Description}
For multi-category 3D anomaly detection, the available data in the training phase contains point cloud samples from multiple categories, i.e., $P_{train}=\left\{P_{train}^1, P_{train}^2,\cdots, P_{train}^c\right\}$, where $P_{train}^i$ denotes the training data from the $i$-th category and only have normal samples, and $c$ is the number of categories. In the testing phase, the data to be detected includes both normal and anomalous point cloud samples from different categories, i.e., $P_{test}=\left\{P_{test}^1, P_{test}^2, \cdots, P_{test}^c\right\}$. The objective is to train a unified model for multiple categories using only normal training data.

\begin{figure*}[ht]
    \centering
    \includegraphics[width=\linewidth]{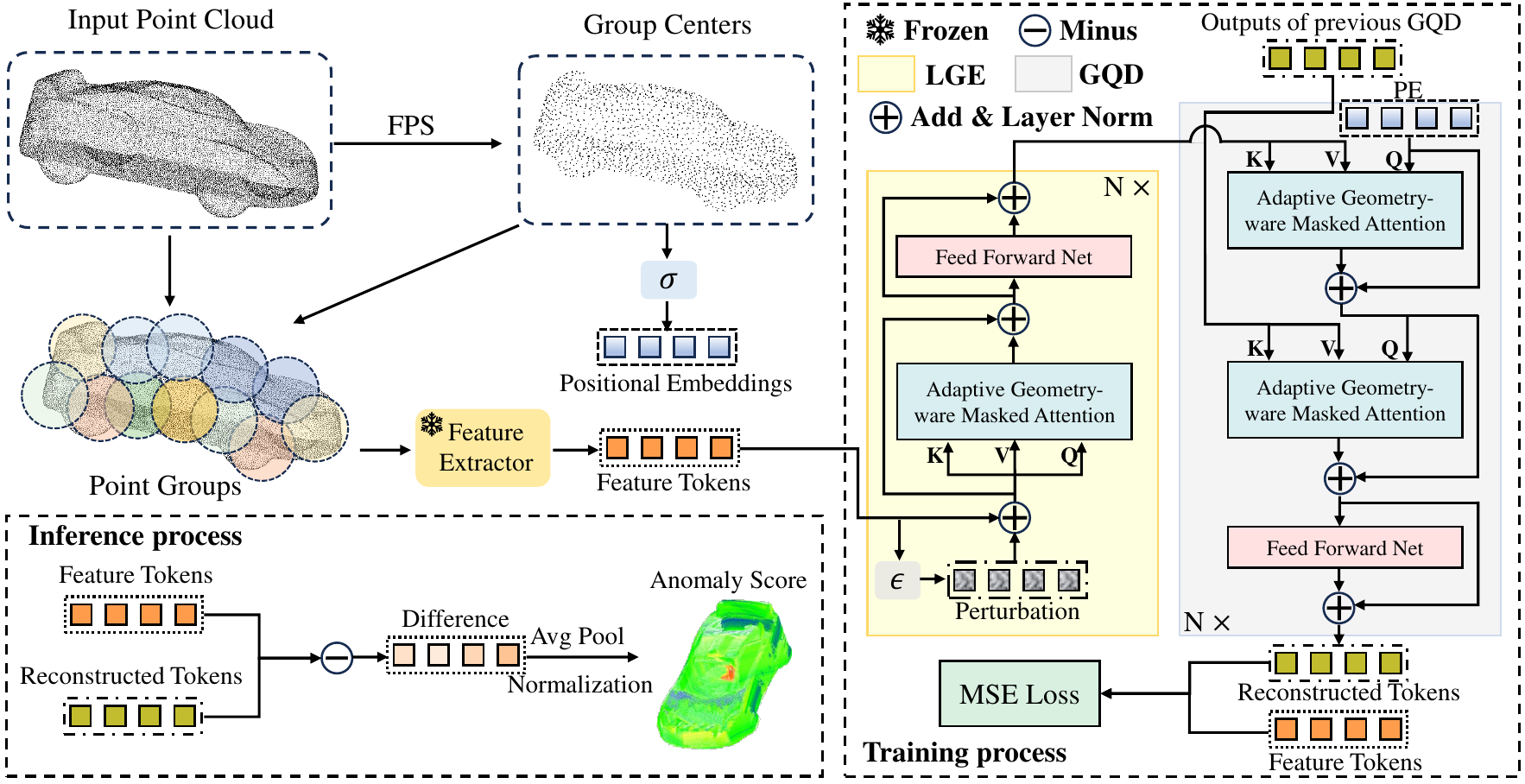}
    \caption{The overview of the proposed method. The input point cloud sample is first registered and aggregated into point groups, while the group centers are projected to obtain position embeddings, and point groups are fed into the feature extractor to generate feature tokens. Then, these tokens and position embeddings are input into the local and global geometry-aware reconstruction framework strengthened by the Adaptive Geometry-aware Masked Attention (AGMA). Finally, anomalies are detected by comparing the differences between the reconstructed and original feature tokens.}
    \label{Overview}
\end{figure*}

\subsection{Overview Framework}
A key challenge in multi-category 3D anomaly detection is to develop a unified representation method to simultaneously adapt to different categories. To this end, we propose a novel geometry-aware reconstruction framework for multi-category 3D anomaly detection. The overall framework is presented in Figure~\ref{Overview}, which consists of three main components: Adaptive Geometry-aware Masked Attention (AGMA), Local Geometry-aware Encoder (LGE), and Global Query Decoder (GQD). Each component is described in the following sections. 

\subsection{Adaptive Geometry-Aware Masked Attention}
The representation of normal point cloud data is the key factor for the success of 3D anomaly detection. Existing reconstruction-based methods use the mask attention mechanism to improve the representation ability. Nevertheless, due to the high variation and complexity of point cloud data, they may fail to learn the intrinsic features to represent normal point cloud samples, leading to the phenomenon of ``identity shortcut". To address this problem, we propose an AGMA module, which explicitly computes neighborhood geometry information for representation, providing not only better reconstruction ability but also model interpretability. 

Specifically, given a point cloud $P=\{p_1, p_2, \cdots, p_n\}$ with each $p_i \in \mathbb{R}^3$, the group centers $\overline{P}_\text{center} = \{\overline{p}_1, \overline{p}_2, \cdots, \overline{p}_m\}$ can be obtained by sampling the point cloud through the Furthest Point Sampling (FPS), which can be expressed as:
\begin{equation}
    \label{eq1}
    \overline{P}_\text{center}=\text{FPS}(P).
\end{equation}

To capture the local geometric information hidden in neighborhood structure, the adaptive neighborhood of group centers is introduced and can be expressed as:
\begin{equation}
    \label{eq2}
    \mathcal{N}_r(\overline{p}_i)=\{\overline{p}_j\in \overline{P}_\text{center}\mid\|\overline{p}_i - \overline{p}_j\|_2\leq r\},
\end{equation}
where $\|\cdot\|_2$ denotes the 2-norm, and $r$ represents the neighborhood radius. Due to the varying scales of point clouds across different categories, an adaptive radius is employed to ensure that the number of points in the neighborhood of each group center point remains consistent. The calculation of $r$ can be expressed as:
\begin{equation}
    \label{eq3}
    r=\frac{\eta}{|\overline{P}_\text{center}|}\sum_{\overline{p}_i\in \overline{P}_\text{center}}\|\overline{p}_i-\overline{p}^\text{nearest}_i\|_2,
\end{equation}
where $|\cdot|$ denotes the cardinality of a set, $p^\text{nearest}_i$ is the nearest neighbor of the $p_i$, and $\eta$ is a scaling factor to adjust the size of the neighborhood.

Subsequently, to reflect the geometric information within the neighborhood structure, we define the normal vector and curvature for each group center point $\overline{p}_i$, which can be formulated as:
\begin{equation}
    \label{eq4}
    \textbf{N}_i=FEV_\text{min}(\text{Cov}_i),
\end{equation}
\begin{equation}
    \label{eq5}
    C_i=\frac{\lambda_\text{min}}{\sum_{j}^3 \lambda_j},
\end{equation}
where $FEV_\text{min}$ means finding the eigenvector corresponding to the smallest eigenvalue, $\lambda_i$ and $\lambda_\text{min}$ represents the $i$-th and the minimum eigenvalue of the covariance matrix, respectively, and $\text{Cov}_i$ is the covariance matrix of group center point $\overline{p}_i$, which is defined as:
\begin{equation}
    \label{eq6}
    \text{Cov}_i=\frac{1}{|\mathcal{N}_r(\overline{p}_i)|}\sum_{\overline{p}_i \in \mathcal{N}_r(\overline{p}_i)}(\overline{p}_i-\mu_i)(\overline{p}_i-\mu_i)^T,
\end{equation}
where $\mu_i$ is the centroid of the neighborhood $\mathcal{N}_r(\overline{p}_i)$ of $\overline{p}_i$. Based on the calculated normal vector $\textbf{N}_i \in \mathbb{R}^3$ and curvature $C_i$, we can further define an index to quantify the variation in geometric information: 
\begin{equation}
    \label{eq7}
    Var^\text{geom}_i=\alpha Var^\text{norm}_i + \beta Var^\text{cruv}_i,
\end{equation}
\begin{equation}
    \label{eq8}
    Var^\text{norm}_i=\frac{1}{|\mathcal{N}_r(\overline{p}_i)|} \sum_{\overline{p}_j \in \mathcal{N}_r(\overline{p}_i)} \angle(\textbf{N}_i,\textbf{N}_j),
\end{equation}
\begin{equation}
    \label{eq9}
    Var^\text{cruv}_i=\frac{1}{|\mathcal{N}_r(\overline{p}_i)|} \sum_{\overline{p}_j \in \mathcal{N}_r(\overline{p}_i)} |C_i-C_j|,
\end{equation}
where $\angle(\cdot,\cdot)$ denotes the angle between two vectors, $Var^\text{norm}_i$, $Var^\text{cruv}_i$, and $Var^\text{geom}_i$ represent the degree of change in the normal vector, curvature, and geometric information within the adaptive neighborhood, respectively, and the $\alpha$ and $\beta$ are hyper-parameters to balance the values.

\begin{figure}[h]
    \centering
    \includegraphics[width=\linewidth]{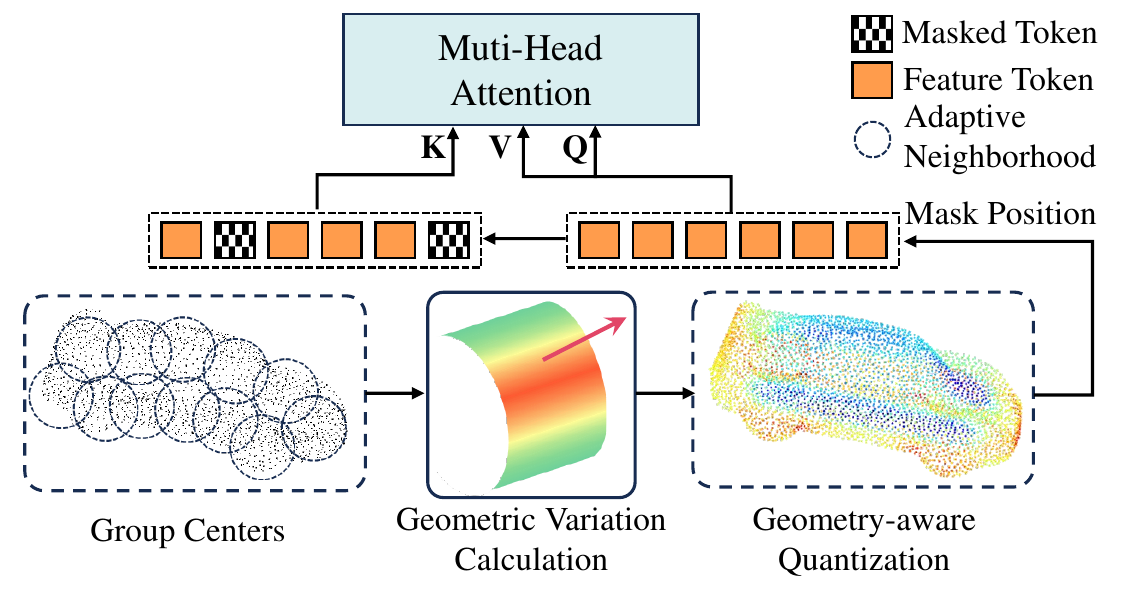}
    \caption{The pipeline of AGMA. The AGMA explicitly extracts geometric variation information from the group center points for mask attention, thereby enhancing the reconstruction representation ability and interpretability.}
    \label{AGMA}
\end{figure}

Intuitively, this geometric change information can be used to guide the learning of representation for reconstruction. Therefore, an improved mask attention mechanism shown in Figure~\ref{AGMA} is introduced, aiming to mask some key feature tokens to enhance representation ability.

Specifically, geometric variation information is first extracted for group center points. Then, points with larger and smaller values are randomly selected, respectively, according to the ratio $\rho$ of tokens to be masked, and their corresponding group feature tokens are masked during attention calculation. With this mask attention mechanism, feature representation ability can be significantly improved, and the interpretability of the model is accordingly enhanced.

\subsection{Local Geometry-Aware Encoder}
Since anomalies usually account for a small area of the entire point cloud, thus encoding local features is essential for improving anomaly detection performance. To encode local group features with geometric information, we propose an LGE to incorporate neighborhood geometric variation information into the feature encoding process. Specifically, a point group is formed by applying $k$-Nearest Neighbors (KNN) to each group center point $\overline{p}_i$, which can be defined as:
\begin{equation}
    G_i=KNN(\overline{p}_i, P).
\end{equation}

This operation is similar to patch extraction in 2D images, aiming to extract local features from data sequences. Subsequently, the point groups $G=\{G_1, G_2,\cdots, G_g\}$ are input into the feature extractor $\mathcal{F}$ to generate group-level feature tokens $\boldsymbol{F_{\text{tok}}} \in \mathbb{R}^{g \times c}$, where $g$ is the number of groups and $c$ is the number of channels. Concurrently, the group centers $P_{\text{center}}$ are passed through a MultiLayer Perceptron (MLP) $\sigma$ to obtain position embeddings $\boldsymbol{F}_{\text{pos}} \in \mathbb{R}^{g \times c}$, which serve as the Transformer's positional encodings. Additionally, before inputting the feature tokens $\boldsymbol{F_{\text{tok}}}$ into the LGE, feature jittering\cite{uniad,noise} is adopted by adding perturbation to the features. This promotes the representation and reconstruction capability through the denoising process. The perturbation for the $i$-th group feature can be expressed as:
\begin{equation}
    \epsilon^i\sim N\big(0, (\gamma\frac{||\boldsymbol{F^i}_{tok}||_2}{C})^2\big),
\end{equation}
where $N(\cdot, \cdot)$ denotes the normal distribution, and $\gamma$ is a scaling factor to control the intensity of perturbation on the group feature token.

Finally, the group feature tokens with added noise are fed into the LGE for local feature encoding. The LGE is composed of $\text{N}$ sequential blocks, each consisting of an AGMA module and a Feed Forward Network (FNN), where the FNN is implemented by a $2$-layer fully connected MLP. By incorporating the AGMA module, the LGE module can encode robust local features with geometric awareness, facilitating subsequent decoding and reconstruction.

\subsection{Global Query Decoder}
Effectively encoding local features only does not ensure accurate and complete anomaly localization. Moreover, it is evident that global information can positively guide the reconstruction process and improve decoding ability. To this end, we propose a GQD module, which leverages global queries to improve anomaly localization. Specifically, the previously obtained position embeddings $\boldsymbol{F}_{\text{pos}}$ are considered as global queries to be fed into an AGMA. Subsequently, the results are further added with the position embeddings again and fed into another AGMA, followed by an FNN as in the LGE. 

To improve the decoding and reconstruction ability, the GQD stacks $\text{N}$ repeated blocks. In each block, the local encoding features from the LGE and the position embeddings acted as global queries are input into the first AGMA. Then, the output of the previous block is combined with the local-global features obtained from the first AGMA and fed into the second AGMA. This interaction promotes feature fusion between blocks and facilitates feature decoding and reconstruction. Finally, the GQD outputs the reconstructed feature tokens $\boldsymbol{F}_{\text{rec}} \in \mathbb{R}^{g \times c}$, which is optimized by the MSE loss:

\begin{table*}[!ht]
\resizebox{\textwidth}{!}{
\begin{tabular}{@{}lcccccccccccc|c@{}}
\toprule
\multicolumn{14}{c}{\textbf{(a) O-AUROC($\uparrow$)}}                                                                                \\ \midrule
\textbf{Method}              & \textbf{Airplane}                     & \textbf{Car}                          & \textbf{Candybar}                     & \textbf{Chicken}                      & \textbf{Diamond}                      & \textbf{Duck}                         & \textbf{Fish}                         & \textbf{Gemstone}                     & \textbf{Seahorse}                     & \textbf{Shell}                        & \textbf{Starfish}                     & \textbf{Toffees}                      & \textbf{Average}                      \\ \midrule
\textbf{BTF(Raw)}            & 0.520                                 & 0.560                                 & 0.462                                 & 0.432                                 & 0.545                                 &  0.784          & 0.549                                 & 0.648                                 & 0.779                                 &  0.754          & 0.575                                 & 0.630                                 & 0.603                                 \\
\textbf{BTF(FPFH)}           & 0.730                                 & 0.647                                 & 0.703                                 & 0.789                                 & 0.707                                 & 0.691                                 & 0.602                                 & {\color[HTML]{FF0000} \textbf{0.686}} & 0.596                                 & 0.396                                 & 0.530                                 & 0.539                                 & 0.635                                 \\
\textbf{M3DM(PointBERT)}     & 0.407                                 & 0.506                                 & 0.442                                 & 0.673                                 & 0.627                                 & 0.466                                 & 0.556                                 & 0.617                                 & 0.494                                 & 0.577                                 & 0.528                                 & 0.562                                 & 0.538                                 \\
\textbf{M3DM(PointMAE)}      & 0.434                                 & 0.541                                 & 0.450                                 & 0.683                                 & 0.602                                 & 0.433                                 & 0.540                                 & 0.644                                 & 0.495                                 & 0.694                                 & 0.551                                 & 0.552                                 & 0.552                                 \\
\textbf{PatchCore(FPFH)}     & {\color[HTML]{FF0000} \textbf{0.882}} & 0.590                                 & 0.565                                 & 0.837                                 & 0.574                                 & 0.546                                 & 0.675                                 & 0.370                                 & 0.505                                 & 0.589                                 & 0.441                                 & 0.541                                 & 0.593                                 \\
\textbf{PatchCore(FPFH+Raw)} & 0.848                                 & {\color[HTML]{FF0000} \textbf{0.777}} & 0.626                                 & {\color[HTML]{FF0000} \textbf{0.853}} & 0.784                                 & 0.628                                 & 0.837                                 & 0.359                                 & 0.767                                 & 0.663                                 & 0.471                                 & 0.570                                 & 0.682                                 \\
\textbf{PatchCore(PointMAE)} & 0.726                                 & 0.498                                 & 0.585                                 & 0.827                                 & 0.783                                 & 0.489                                 & 0.630                                 & 0.374                                 & 0.539                                 & 0.501                                 & 0.519                                 & 0.663                                 & 0.594                                 \\
\textbf{CPMF}                & 0.632                                 & 0.518                                 & 0.718                                 & 0.640                                 & 0.640                                 & 0.554                                 & 0.840                                 & 0.349                                 & {\color[HTML]{FF0000} \textbf{0.843}} & 0.393                                 & 0.526                                 & {\color[HTML]{FF0000} \textbf{0.845}} & 0.625                                 \\
\textbf{IMRNet}              & 0.762                                 & 0.711                                 & 0.755                                 & 0.780                                 & 0.905                                 & 0.517                                 & 0.880                                 & {\color[HTML]{0070C0} \textbf{0.674}} & 0.604                                 & 0.665                                 &  0.674          & 0.774                                 & 0.725                                 \\
\textbf{Reg3D-AD}            & 0.716                                 & 0.697                                 & 0.827                                 & {\color[HTML]{0070C0} \textbf{0.852}} & 0.900                                 & 0.584                                 & {\color[HTML]{0070C0} \textbf{0.915}} & 0.417                                 & 0.762                                 & 0.583                                 & 0.506                                 & 0.685                                 & 0.704                                 \\
\textbf{Group3AD}            & 0.744                                 & 0.728                                 & {\color[HTML]{FF0000} \textbf{0.847}} & 0.786                                 & {\color[HTML]{0070C0} \textbf{0.932}} & 0.679                                 & {\color[HTML]{FF0000} \textbf{0.976}} & 0.539                                 & {\color[HTML]{0070C0} \textbf{0.841}} & 0.585                                 & 0.562                                 & {\color[HTML]{0070C0} \textbf{0.796}} & {\color[HTML]{0070C0} \textbf{0.751}} \\
\textbf{R3D-AD}              & 0.772                                 & 0.696                                 & 0.713                                 & 0.714                                 & 0.685                                 & {\color[HTML]{FF0000} \textbf{0.909}} & 0.692                                 & 0.665                                 & 0.720                                 & {\color[HTML]{FF0000} \textbf{0.840}} & {\color[HTML]{0070C0} \textbf{0.701}} & 0.703                                 & 0.734                                 \\
\textbf{Ours}                & {\color[HTML]{0070C0} \textbf{0.850}} & {\color[HTML]{0070C0} \textbf{0.749}} & {\color[HTML]{0070C0} \textbf{0.830}} & 0.715                                 & {\color[HTML]{FF0000} \textbf{0.955}} & {\color[HTML]{0070C0} \textbf{0.831}} & 0.865                                 & 0.560                                 & 0.716                                 & {\color[HTML]{0070C0} \textbf{0.803}} & {\color[HTML]{FF0000} \textbf{0.766}} & 0.738                                 & {\color[HTML]{FF0000} \textbf{0.782}} \\ \midrule
                             & \multicolumn{1}{l}{}                  & \multicolumn{1}{l}{}                  & \multicolumn{1}{l}{}                  & \multicolumn{1}{l}{}                  & \multicolumn{1}{l}{}                  & \multicolumn{1}{l}{}                  & \multicolumn{1}{l}{}                  & \multicolumn{1}{l}{}                  & \multicolumn{1}{l}{}                  & \multicolumn{1}{l}{}                  & \multicolumn{1}{l}{}                  & \multicolumn{1}{l}{}                  & \multicolumn{1}{l}{}                  \\ \midrule
\multicolumn{14}{c}{\textbf{(b) P-AUROC($\uparrow$)}}                                                                                                                                                                                                                                       \\ \midrule
\textbf{Method}              & \textbf{Airplane}                     & \textbf{Car}                          & \textbf{Candybar}                     & \textbf{Chicken}                      & \textbf{Diamond}                      & \textbf{Duck}                         & \textbf{Fish}                         & \textbf{Gemstone}                     & \textbf{Seahorse}                     & \textbf{Shell}                        & \textbf{Starfish}                     & \textbf{Toffees}                      & \textbf{Average}                      \\ \midrule
\textbf{BTF(Raw)}            & 0.564                                 & 0.647                                 & 0.735                                 & 0.608                                 & 0.563                                 & 0.601                                 & 0.514                                 & 0.597                                 & 0.520                                 & 0.489 & 0.392                                 & 0.623                                 & 0.571                                 \\
\textbf{BTF(FPFH)}           & {\color[HTML]{FF0000} \textbf{0.738}} & 0.708                                 & {\color[HTML]{0070C0} \textbf{0.864}} & 0.693                                 & {\color[HTML]{0070C0} \textbf{0.882}}                                 & {\color[HTML]{FF0000} \textbf{0.875}} & 0.709                                 & {\color[HTML]{FF0000} \textbf{0.891}} & 0.512                                 & 0.571                                 & 0.501                                 & 0.815                                 & 0.730                                 \\
\textbf{M3DM(PointBERT)}     & 0.523                                 & 0.593                                 & 0.682                                 & {\color[HTML]{FF0000} \textbf{0.790}} & 0.594                                 & 0.668                                 & 0.589                                 & 0.646                                 & 0.574                                 & 0.732                                 & 0.563                                 & 0.677                                 & 0.636                                 \\
\textbf{M3DM(PointMAE)}      & 0.530                                 & 0.607                                 & 0.683                                 & 0.735                                 & 0.618                                 & 0.678                                 & 0.600                                 & 0.654                                 & 0.561                                 & 0.748                                 & 0.555                                 & 0.679                                 & 0.637                                 \\
\textbf{PatchCore(FPFH)}     & 0.471                                 & 0.643                                 & 0.637                                 & 0.618                                 & 0.760                                 & 0.430                                 & 0.464                                 & {\color[HTML]{0070C0} \textbf{0.830}} & 0.544                                 & 0.596                                 & 0.522                                 & 0.411                                 & 0.577                                 \\
\textbf{PatchCore(FPFH+Raw)} & 0.556                                 & 0.740                                 & 0.749                                 & 0.558                                 & 0.854                                 & 0.658                                 & 0.781                                 & 0.539                                 & 0.808                                 & 0.753                                 & 0.613                                 & 0.549                                 & 0.680                                 \\
\textbf{PatchCore(PointMAE)} & 0.579                                 & 0.610                                 & 0.635                                 & 0.683                                 & 0.776                                 & 0.439                                 & 0.714                                 & 0.514                                 & 0.660                                 & 0.725                                 & 0.641                                 & 0.727                                 & 0.642                                 \\
\textbf{CPMF}                & 0.618                                 & {\color[HTML]{FF0000} \textbf{0.836}} & 0.734                                 & 0.559                                 & 0.753                                 & 0.719                                 & {\color[HTML]{FF0000} \textbf{0.988}} & 0.449                                 & {\color[HTML]{FF0000} \textbf{0.962}} & 0.725                                 & {\color[HTML]{FF0000} \textbf{0.800}} & {\color[HTML]{FF0000} \textbf{0.959}} & {\color[HTML]{0070C0} \textbf{0.758}} \\
\textbf{Reg3D-AD}            & 0.631                                 & 0.718                                 & 0.724                                 & 0.676 & 0.835                                 & 0.503                                 & 0.826                                 & 0.545                                 & 0.817                                 & {\color[HTML]{FF0000} \textbf{0.811}}                                 & 0.617                                 & 0.759                                 & 0.705                                 \\
\textbf{Group3AD}            & {\color[HTML]{0070C0} \textbf{0.636}} & 0.745                                 & 0.738                                 & {\color[HTML]{0070C0} \textbf{0.759}}                                 & 0.862 & 0.631                                 & 0.836 & 0.564                                 & {\color[HTML]{0070C0} \textbf{0.827}} & {\color[HTML]{0070C0} \textbf{0.798}}                                 & 0.625                                 & 0.803                                 & 0.735                                 \\
\textbf{Ours}                & 0.628                                 & {\color[HTML]{0070C0} \textbf{0.819}} & {\color[HTML]{FF0000} \textbf{0.910}} & 0.640                                 & {\color[HTML]{FF0000} \textbf{0.942}} & {\color[HTML]{0070C0} \textbf{0.822}} & {\color[HTML]{0070C0} \textbf{0.932}}                                 & 0.458                                 & 0.659                                 & 0.778 & {\color[HTML]{0070C0} \textbf{0.690}} & {\color[HTML]{0070C0} \textbf{0.934}} & {\color[HTML]{FF0000} \textbf{0.768}} \\ \bottomrule
\end{tabular}}
\caption{The experimental results for anomaly detection across 12 categories of Real3D-AD. The best and the second-best results are highlighted in \textcolor[rgb]{1, 0, 0}{\textbf{red}}  and {\color[HTML]{0070C0}\textbf{blue}}, respectively. The results of the baselines are excerpted from their papers.}
  \label{Real3D}
\end{table*}
\begin{equation}
    \mathcal{L}_=\frac{1}{g}\sum_{i=1}^g\|{\boldsymbol{F}}^i_{\text{tok}}-\boldsymbol{F}^i_{\text{rec}}\|_2.
\end{equation}

During the testing phase, the test point cloud is first registered and grouped for the feature extractor to generate feature tokens. Then, the proposed reconstruction model tries to encode and decode them into the original form. The reconstruction difference is normalized and subjected to average pooling to obtain the final pixel-level anomaly score $S_p$, which can be expressed as:
\begin{equation}
    S_p = AvgPool\big(Norm(\parallel \boldsymbol{F}_{\text{rec}}- \boldsymbol{F_{\text{tok}}}\parallel_2)\big),
\end{equation}
where $Norm$ denotes the min-max normalization, and $AvgPool$ means the operation of average pooling with the kernel size of 1*512. The anomaly score indicates the likelihood of the point being anomalous, and the maximum value of $S_p$ is used as the object-level anomaly score $S_o$.

\section{Experiment}
\subsection{Experiment Settings}
\textbf{Datasets.}
(1) Real3D-AD~\cite{real3dad} is a high-resolution point cloud anomaly detection dataset consisting of 1,254 samples from 12 object categories. Each category has only four training samples but contains anomalies with varying shapes and sizes.
(2) Anomaly-ShapeNet~\cite{IMRNet} is a synthetic point cloud anomaly detection dataset containing 1,600 samples across 40 categories. Each sample contains between 8,000 and 30,000 points, with the anomalous region accounting for 1\% to 10\% of the entire point cloud. Due to the large number of categories, this dataset is more challenging for multi-class anomaly detection.

\textbf{Comparison Methods.} Our method adopts the setting of multi-category anomaly detection, where only one model is uniformly trained for all categories. Because existing methods can not apply to multiple categories directly, the compared methods in the experiment use the single-category configuration, wherein privately owned models are separately trained for each category. The proposed method is compared with some representative methods, including BTF~\cite{Horwitz2023BTF}, M3DM~\cite{Wang2023multimodal}, PatchCore\cite{patchcore}, CMPF~\cite{cao2023complementarypseudomultimodalfeature}, Reg3D-AD\cite{real3dad}, Group3AD~\cite{zhu2024towards}, IMRNet~\cite{IMRNet}, and R3D-AD~\cite{Zhou2024R3D}.

\textbf{Implementation Details.} PointMAE pre-trained on ModelNet408K~\cite{modelnet40} is adopted as the feature extractor of our method. The AdamW optimizer is used in the training process, and the learning rate is initially set to 0.0001 and dropped to 0.00001 after 800 epochs. The batch size and the maximum number of epochs are set to 1 and 1000, respectively. The hyperparameters $\alpha$, $\beta$, $\eta$, and $\rho$ for AGMA are set to 1, 10, 7, and 0.4, respectively. The number of stacked blocks $\text{N}$ in LGE and GQD is set to 4. Our method is performed on PyTorch 1.13.0 and CUDA 11.7 with an NVIDIA A100-PCIE-40GB GPU.

\textbf{Evaluation Metrics.} In the experiments, the Area under the Receiver Operating Characteristic Curve (AUROC, ↑) is used to assess the performance of object-level anomaly detection and pixel-level anomaly localization.
\begin{table*}[t!]
  \centering
\resizebox{\textwidth}{!}{

    \begin{tabular}{l|cccccccccccccc}
    \toprule
    \multicolumn{15}{c}{\textbf{O-AUROC($\uparrow$)}} \\
    \midrule
    \textbf{Method} & \textbf{cap0} & \textbf{cap3} & \textbf{helmet3} & \textbf{cup0} & \textbf{bowl4} & \textbf{vase3} & \textbf{headset1} & \textbf{eraser0} & \textbf{vase8} & \textbf{cap4} & \textbf{vase2} & \textbf{vase4} & \textbf{helmet0} & \textbf{bucket1} \\
    \midrule
    \textbf{BTF(Raw)}            & 0.668                                 & 0.527                                 & 0.526                                 & 0.403                                 & 0.664                                 & 0.717                                 & 0.515                                 & 0.525                                 & 0.424                                 & 0.468                                 & 0.410                                 & 0.425                                 & 0.553                                 & 0.321                                 \\
\textbf{BTF(FPFH)}           & 0.618                                 & 0.522                                 & 0.444                                 & 0.586                                 & 0.609                                 & 0.699                                 & 0.490                                 & 0.719                                 & 0.668                                 & 0.520                                 & 0.546                                 & 0.510                                 & 0.571                                 & 0.633                                 \\
\textbf{M3DM}                & 0.557                                 & 0.423                                 & 0.374                                 & 0.539                                 & 0.464                                 & 0.439                                 & 0.617                                 & 0.627                                 & 0.663                                 & {\color[HTML]{0070C0} \textbf{0.777}} & 0.737                                 & 0.476                                 & 0.526                                 & 0.501                                 \\
\textbf{Patchcore(FPFH)}     & 0.580                                 & 0.453                                 & 0.404                                 & 0.600                                 & 0.494                                 & 0.449                                 & 0.637                                 & 0.657                                 & 0.662                                 & 0.757                                 & 0.721                                 & 0.506                                 & 0.546                                 & 0.551                                 \\
\textbf{Patchcore(PointMAE)} & 0.589                                 & 0.476                                 & 0.424                                 & 0.610                                 & 0.501                                 & 0.460                                 & 0.627                                 & 0.677                                 & 0.663                                 & 0.727                                 & 0.741                                 & 0.516                                 & 0.556                                 & 0.561                                 \\
\textbf{CPMF}                & 0.601                                 & 0.551                                 & 0.520                                 & 0.497                                 & 0.683                                 & 0.582                                 & 0.458                                 & 0.689                                 & 0.529                                 & 0.553                                 & 0.582                                 & 0.514                                 & 0.555                                 & 0.601                                 \\
\textbf{Reg3D-AD}               & 0.693                                 & 0.725                                 & 0.367                                 & 0.510                                 & 0.663                                 & 0.650                                 & 0.610                                 & 0.343                                 & 0.620                                 & 0.643                                 & 0.605                                 & 0.500                                 & 0.600                                 & 0.752                                 \\
\textbf{IMRNet}              & 0.737                                 & {\color[HTML]{FF0000} \textbf{0.775}} & 0.573                                 & 0.643                                 & 0.676                                 & 0.700                                 & 0.676                                 & 0.548                                 & 0.630                                 & 0.652                                 & 0.614                                 & 0.524                                 & 0.597                                 & {\color[HTML]{0070C0} \textbf{0.771}} \\
\textbf{R3D-AD}              & {\color[HTML]{FF0000} \textbf{0.822}} & {\color[HTML]{0070C0} \textbf{0.730}} & {\color[HTML]{0070C0} \textbf{0.707}} & {\color[HTML]{FF0000} \textbf{0.776}} & {\color[HTML]{0070C0} \textbf{0.744}} & {\color[HTML]{0070C0} \textbf{0.742}} & {\color[HTML]{0070C0} \textbf{0.795}} & {\color[HTML]{FF0000} \textbf{0.890}} & {\color[HTML]{FF0000} \textbf{0.721}} & 0.681                                 & {\color[HTML]{0070C0} \textbf{0.752}} & {\color[HTML]{0070C0} \textbf{0.630}} & {\color[HTML]{FF0000} \textbf{0.757}} & 0.756                                 \\
\textbf{Ours}                & {\color[HTML]{0070C0} \textbf{0.793}} & 0.701                                 & {\color[HTML]{FF0000} \textbf{0.979}} & {\color[HTML]{0070C0} \textbf{0.743}} & {\color[HTML]{FF0000} \textbf{0.911}} & {\color[HTML]{FF0000} \textbf{0.761}} & {\color[HTML]{FF0000} \textbf{0.886}} & {\color[HTML]{0070C0} \textbf{0.776}} & {\color[HTML]{0070C0} \textbf{0.670}} & {\color[HTML]{FF0000} \textbf{0.835}} & {\color[HTML]{FF0000} \textbf{0.929}} & {\color[HTML]{FF0000} \textbf{0.876}} & {\color[HTML]{0070C0} \textbf{0.672}} & {\color[HTML]{FF0000} \textbf{0.784}} \\
    \midrule
    \multicolumn{1}{c}{} &       &       &       &       &       &       &       &       &       &       &       &       &       &  \\
    \midrule
    \textbf{Method} & \textbf{bottle3} & \textbf{vase0} & \textbf{bottle0} & \textbf{tap1} & \textbf{bowl0} & \textbf{bucket0} & \textbf{vase5} & \textbf{vase1} & \textbf{vase9} & \textbf{ashtray0} & \textbf{bottle1} & \textbf{tap0} & \textbf{phone} & \textbf{cup1} \\
    \midrule
    \textbf{BTF(Raw)}            & 0.568                                 & 0.531                                 & 0.597                                 & 0.573                                 & 0.564                                 & 0.617                                 & 0.585                                 & 0.549                                 & 0.564                                 & 0.578                                 & 0.510                                 & 0.525                                 & 0.563                                 & 0.521                                 \\
\textbf{BTF(FPFH)}           & 0.322                                 & 0.342                                 & 0.344                                 & 0.546                                 & 0.509                                 & 0.401                                 & 0.409                                 & 0.219                                 & 0.268                                 & 0.420                                 & 0.546                                 & 0.560                                 & 0.671                                 & 0.610                                 \\
\textbf{M3DM}                & 0.541                                 & 0.423                                 & 0.574                                 & 0.739                                 & 0.634                                 & 0.309                                 & 0.317                                 & 0.427                                 & 0.663                                 & 0.577                                 & 0.637                                 & {\color[HTML]{0070C0} \textbf{0.754}} & 0.357                                 & 0.556                                 \\
\textbf{Patchcore(FPFH)}     & 0.572                                 & 0.455                                 & 0.604                                 & 0.766                                 & 0.504                                 & 0.469                                 & 0.417                                 & 0.423                                 & 0.660                                 & 0.587                                 & 0.667                                 & 0.753                                 & 0.388                                 & 0.586                                 \\
\textbf{Patchcore(PointMAE)} & 0.650                                 & 0.447                                 & 0.513                                 & 0.538                                 & 0.523                                 & 0.593                                 & 0.579                                 & 0.552                                 & 0.629                                 & 0.591                                 & 0.601                                 & 0.458                                 & 0.488                                 & 0.556                                 \\
\textbf{CPMF}                & 0.405                                 & 0.451                                 & 0.520                                 & 0.697                                 & 0.783                                 & 0.482                                 & 0.618                                 & 0.345                                 & 0.609                                 & 0.353                                 & 0.482                                 & 0.359                                 & 0.509                                 & 0.499                                 \\
\textbf{Reg3D-AD}               & 0.525                                 & 0.533                                 & 0.486                                 & 0.641                                 & 0.671                                 & 0.610                                 & 0.520                                 & 0.702                                 & 0.594                                 & 0.597                                 & 0.695                                 & 0.676                                 & 0.414                                 & 0.538                                 \\
\textbf{IMRNet}              & 0.640                                 & 0.533                                 & 0.552                                 & 0.696                                 & 0.681                                 & 0.580                                 & 0.676                                 & {\color[HTML]{0070C0} \textbf{0.757}} & 0.594                                 & 0.671                                 & 0.700                                 & 0.676                                 & 0.755                                 & {\color[HTML]{0070C0} \textbf{0.757}} \\
\textbf{R3D-AD}              & {\color[HTML]{FF0000} \textbf{0.781}} & {\color[HTML]{0070C0} \textbf{0.788}} & {\color[HTML]{0070C0} \textbf{0.733}} & {\color[HTML]{0070C0} \textbf{0.900}} & {\color[HTML]{0070C0} \textbf{0.819}} & {\color[HTML]{0070C0} \textbf{0.683}} & {\color[HTML]{0070C0} \textbf{0.757}} & 0.729                                 & {\color[HTML]{0070C0} \textbf{0.718}} & {\color[HTML]{0070C0} \textbf{0.833}} & {\color[HTML]{FF0000} \textbf{0.737}} & 0.736                                 & {\color[HTML]{0070C0} \textbf{0.762}} & {\color[HTML]{0070C0} \textbf{0.757}} \\
\textbf{Ours}                & {\color[HTML]{0070C0} \textbf{0.756}} & {\color[HTML]{FF0000} \textbf{0.821}} & {\color[HTML]{FF0000} \textbf{0.795}} & {\color[HTML]{FF0000} \textbf{0.970}} & {\color[HTML]{FF0000} \textbf{0.930}} & {\color[HTML]{FF0000} \textbf{0.898}} & {\color[HTML]{FF0000} \textbf{0.976}} & {\color[HTML]{FF0000} \textbf{0.857}} & {\color[HTML]{FF0000} \textbf{0.736}} & {\color[HTML]{FF0000} \textbf{0.962}} & {\color[HTML]{0070C0} \textbf{0.709}} & {\color[HTML]{FF0000} \textbf{0.945}} & {\color[HTML]{FF0000} \textbf{0.919}} & {\color[HTML]{FF0000} \textbf{0.952}} \\
    \midrule
    \multicolumn{1}{c}{} &       &       &       &       &       &       &       &       &       &       &       &       &       &  \\
    \midrule
    \textbf{Method} & \textbf{vase7} & \textbf{helmet2} & \textbf{cap5} & \textbf{shelf0} & \textbf{bowl5} & \textbf{bowl3} & \textbf{helmet1} & \textbf{bowl1} & \textbf{headset0} & \textbf{bag0} & \textbf{bowl2} & \textbf{jar} & \multicolumn{2}{|c}{\textbf{Mean}} \\
    \midrule
    \textbf{BTF(Raw)}            & 0.448                                 & 0.602                                 & 0.373                                 & 0.164                                 & 0.417                                 & 0.385                                 & 0.349                                 & 0.264                                 & 0.378                                 & 0.410                                 & 0.525                                 & \multicolumn{1}{c|}{0.420}                                 & \multicolumn{2}{c}{0.493}                                 \\
\textbf{BTF(FPFH)}           & 0.518                                 & 0.542                                 & 0.586                                 & 0.609                                 & 0.699                                 & 0.490                                 & 0.719                                 & 0.668                                 & 0.520                                 & 0.546                                 & 0.510                                 & \multicolumn{1}{c|}{0.424}                                 & \multicolumn{2}{c}{0.528}                                 \\
\textbf{M3DM}                & 0.657                                 & 0.623                                 & 0.639                                 & 0.564                                 & 0.409                                 & 0.617                                 & 0.427                                 & 0.663                                 & 0.577                                 & 0.537                                 & 0.684                                 & \multicolumn{1}{c|}{0.441}                                 & \multicolumn{2}{c}{0.552}                                 \\
\textbf{Patchcore(FPFH)}     & 0.693                                 & 0.425                                 & {\color[HTML]{FF0000} \textbf{0.790}} & 0.494                                 & 0.558                                 & 0.537                                 & 0.484                                 & 0.639                                 & 0.583                                 & 0.571                                 & 0.615                                 & \multicolumn{1}{c|}{0.472}                                 & \multicolumn{2}{c}{0.568}                                 \\
\textbf{Patchcore(PointMAE)} & 0.650                                 & 0.447                                 & 0.538                                 & 0.523                                 & 0.593                                 & 0.579                                 & 0.552                                 & 0.629                                 & 0.591                                 & 0.601                                 & 0.458                                 & \multicolumn{1}{c|}{0.483}                                 & \multicolumn{2}{c}{0.562}                                 \\
\textbf{CPMF}                & 0.397                                 & 0.462                                 & 0.697                                 & 0.685                                 & 0.685                                 & 0.658                                 & 0.589                                 & 0.639                                 & 0.643                                 & 0.643                                 & 0.625                                 & \multicolumn{1}{c|}{0.610}                                 & \multicolumn{2}{c}{0.559}                                 \\
\textbf{Reg3D-AD}               & 0.462                                 & 0.614                                 & 0.467                                 & 0.688                                 & 0.593                                 & 0.348                                 & 0.381                                 & 0.525                                 & 0.537                                 & 0.706                                 & 0.490                                 & \multicolumn{1}{c|}{0.592}                                 & \multicolumn{2}{c}{0.572}                                 \\
\textbf{IMRNet}              & 0.635                                 & {\color[HTML]{FF0000} \textbf{0.641}} & 0.652                                 & 0.603                                 & {\color[HTML]{0070C0} \textbf{0.710}} & 0.599                                 & 0.600                                 & 0.702                                 & 0.720                                 & 0.660                                 & 0.685                                 & \multicolumn{1}{c|}{0.780}                                 & \multicolumn{2}{c}{0.661}                                 \\
\textbf{R3D-AD}              & {\color[HTML]{0070C0} \textbf{0.771}} & {\color[HTML]{0070C0} \textbf{0.633}} & 0.670                                 & {\color[HTML]{0070C0} \textbf{0.696}} & 0.656                                 & {\color[HTML]{0070C0} \textbf{0.767}} & {\color[HTML]{0070C0} \textbf{0.720}} & {\color[HTML]{0070C0} \textbf{0.778}} & {\color[HTML]{0070C0} \textbf{0.738}} & {\color[HTML]{0070C0} \textbf{0.720}} & {\color[HTML]{FF0000} \textbf{0.741}} & \multicolumn{1}{c|}{{\color[HTML]{0070C0} \textbf{0.838}}} & \multicolumn{2}{c}{{\color[HTML]{0070C0} \textbf{0.749}}} \\
\textbf{Ours}                & {\color[HTML]{FF0000} \textbf{0.938}} & 0.609                                 & {\color[HTML]{0070C0} \textbf{0.761}} & {\color[HTML]{FF0000} \textbf{0.841}} & {\color[HTML]{FF0000} \textbf{0.754}} & {\color[HTML]{FF0000} \textbf{0.885}} & {\color[HTML]{FF0000} \textbf{1.000}}  & {\color[HTML]{FF0000} \textbf{0.978}} & {\color[HTML]{FF0000} \textbf{0.862}} & {\color[HTML]{FF0000} \textbf{0.805}} & {\color[HTML]{0070C0} \textbf{0.719}} & \multicolumn{1}{c|}{{\color[HTML]{FF0000} \textbf{0.971}}} & \multicolumn{2}{c}{{\color[HTML]{FF0000} \textbf{0.842}}} \\
    \bottomrule
    \end{tabular}%
}
  \caption{The object-level AUROC experimental results for anomaly detection across 40 categories of Anomaly-ShapeNet. The best and the second-best results are highlighted in \textcolor[rgb]{1, 0, 0}{\textbf{red}}  and {\color[HTML]{0070C0}{\textbf{blue}}}, respectively. The results of the baselines are excerpted from their papers, and the pixel-level AUROC experimental results are provided in the supplementary material.}
  \label{Anomaly-ShapNet-o}%
\end{table*}

\subsection{Main Results}
\textbf{Results on Real3D-AD.} The comparison results of MC3D-AD and the existing methods are shown in Table~\ref{Real3D}. It can be observed that MC3D-AD achieved SOTA performance, with an AUROC of 0.782 at the object level and 0.768 at the pixel level. These results are 3.1\% and 1.0\% higher than those of the second-best method, which is a task-specific model that carefully tunes for each category.

\textbf{Results on Anomaly-ShapeNet.} The experimental results of MC3D-AD on the Anomaly-ShapeNet dataset are presented in Table~\ref{Anomaly-ShapNet-o}. Despite the increased complexity of this dataset, which comprises 40 categories, MC3D-AD still achieved SOTA performance. Specifically, the object-level AUROC for anomaly detection reached 0.842, outperforming the second-best single-category method by 9.3\%. Additionally, the pixel-level AUROC for anomaly localization attained 0.748, which is 8.0\% higher than the second-best single-category approach. These results clearly demonstrate the generalization capabilities of MC3D-AD in multi-class anomaly detection setting.

\subsection{Ablation Studies}
Table~\ref{ablation} presents the results of the experiments evaluating the effectiveness of the proposed modules. Specifically, LGE refers to a model that incorporates only the LGE module, without the AGMA module. LGE$_{AGMA}$ refers to the application of AGMA within the LGE framework, while GQD$_{AGMA}$ indicates the addition of AGMA to the GQD module. 
The results show that AGMA plays a crucial role in capturing point cloud geometry features, significantly enhancing its reconstruction capability. This leads to a significant increase in anomaly detection performance, with a 9.4\% improvement in O-AUROC. On the other hand, LQD provides additional guidance for the model's reconstruction process, further improving anomaly localization and yielding an 8.3\% improvement in P-AUROC. AGMA can be seamlessly integrated into both LGE and GQD, resulting in an overall performance enhancement across all metrics.
\begin{table}[!ht]
\footnotesize
\centering
\begin{tabular}{@{}lcc@{}}
\toprule
\textbf{Method}               & \textbf{O-AUROC} ($\uparrow$) & \textbf{P-AUROC} ($\uparrow$) \\ \midrule
LGE             & 0.658   & 0.650   \\
LGE$_\text{AGMA}$     & 0.752   & 0.709   \\
LGE+GQD         & 0.741   & 0.733   \\
LGE+GQD$_\text{AGMA}$   & 0.756   & 0.755   \\
LGE$_\text{AGMA}$+GQD & 0.755   & 0.760   \\
\textbf{MC3D-AD }            & \textbf{0.782}   & \textbf{0.768}   \\ \bottomrule
\end{tabular}
\caption{Ablation results on Real3D-AD.}
\label{ablation}
\end{table}

\setcounter{figure}{4}
\begin{figure*}[!ht]
    \centering
    \includegraphics[width=0.8\linewidth]{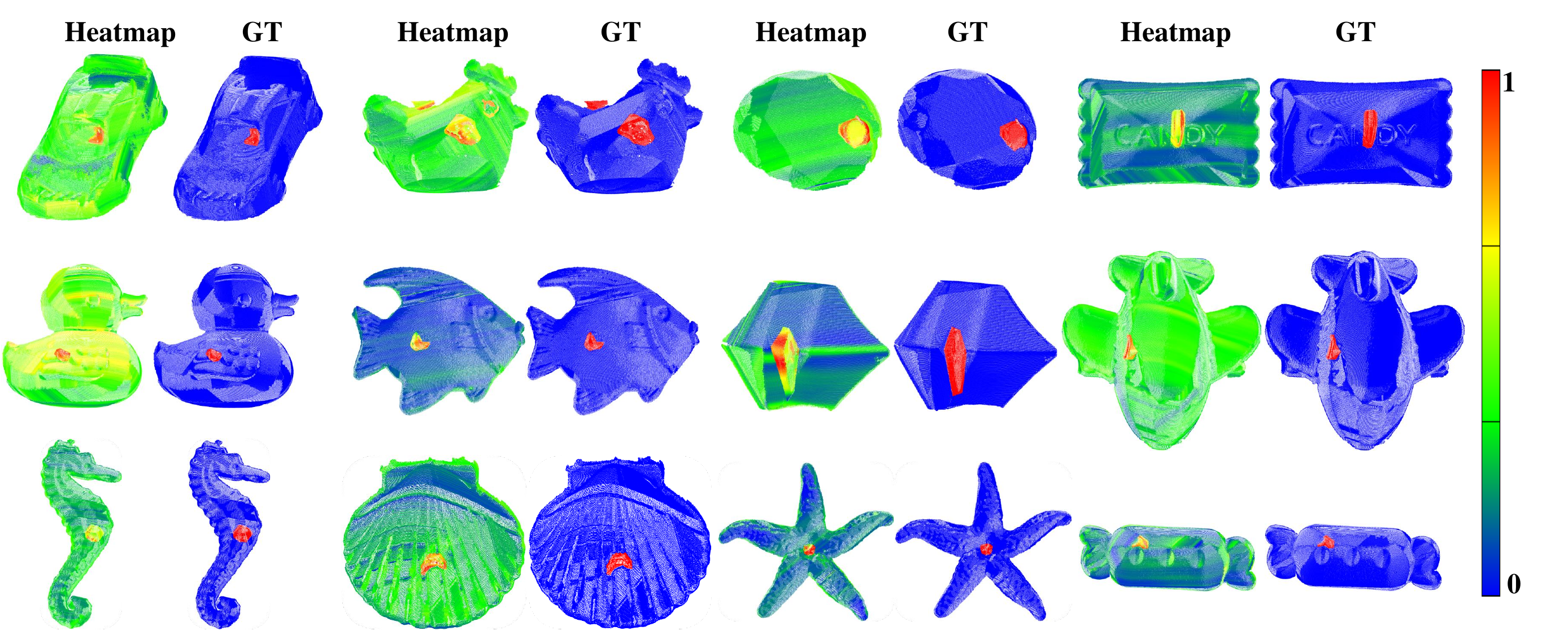}
    \caption{Point heatmap comparison of our MC3D-AD with the Ground Truth (GT) on Real3D-AD. As evidenced by the red-colored areas in the visualized heatmaps, M3DM accurately detects and localizes anomalous regions within the point clouds from different categories.}
    \label{heatmap_vis}
\end{figure*}

\subsection{Analysis of Hyper-Parameters}
Our method introduces two key parameters $\eta$ and $\rho$ to control the size of the adaptive neighborhood and the proportion of masked tokens, respectively. As shown in Figure~\ref{parameters}, an inappropriate $\eta$ value, whether too small or too large, can lead to inaccurate quantization of geometric variation, so a balanced value $\eta=7$ is set to ensure stable performance. Similarly, the mask proportion in the attention mechanism requires careful tuning: a low mask ratio hinders the learning of robust reconstruction, while a high mask ratio increases the difficulty of reconstruction. In the experiments, the parameter $\rho$ is set to 0.4, at which the best performance is reached. 
\setcounter{figure}{3}
\begin{figure}[h!]
    \centering
    \includegraphics[width=\linewidth]{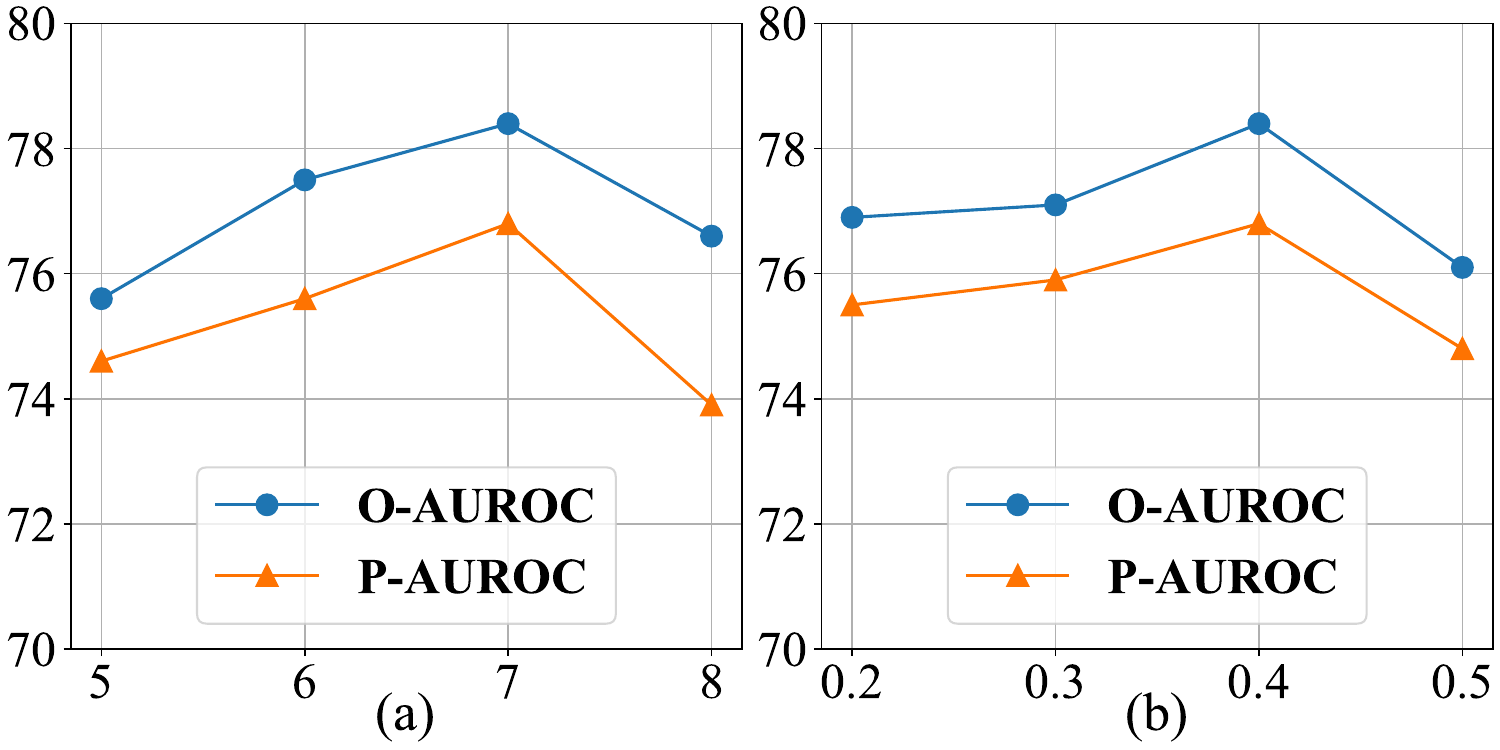}
    \caption{Parameter sensitivity analysis on Real3D-AD. (a) $\eta$; (b) $\rho$.}
    \label{parameters}
\end{figure}

\subsection{Visualization}
Figure~\ref{heatmap_vis} shows the heatmap visualization results of our MC3D-AD on the Real3D-AD. It is observed that MC3D-AD accurately detects and localizes anomalous regions within the point cloud, clearly demonstrating its effectiveness. In addition, Figure~\ref{AGMA_vis} provides the heatmap visualization of the proposed AGMA, which encapsulates the geometric information extracted from the point cloud. The blue regions indicate that the geometric information of the point cloud varies slowly within the adaptive neighborhood, while the transition from green to yellow and red indicates a gradual increase in the change of geometric information within the neighborhood. By intentionally masking blue or red areas during training, the reconstruction ability of our method is greatly improved.
\setcounter{figure}{5}
\begin{figure}[!h]
    \centering
    \includegraphics[width=0.9\linewidth]{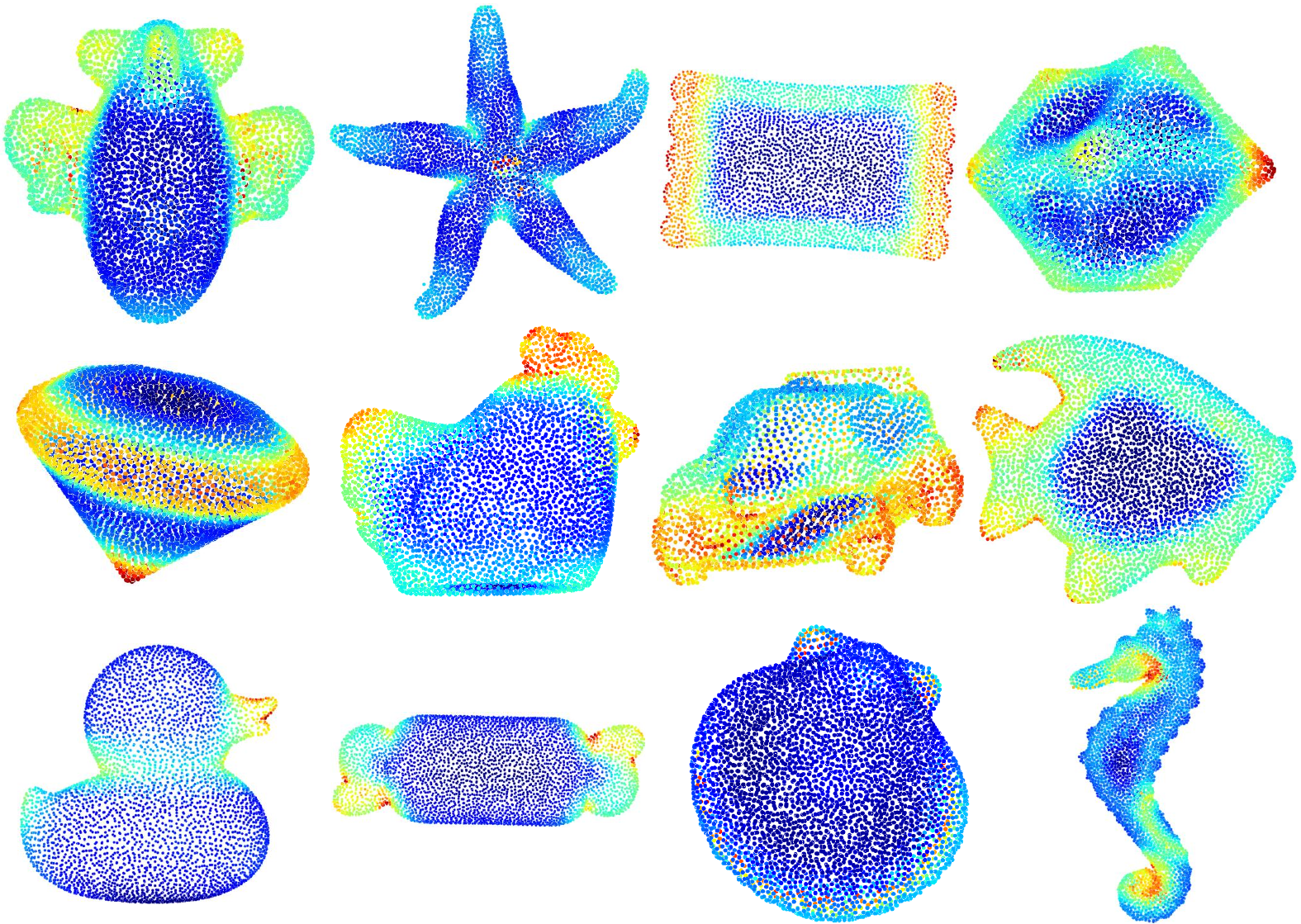}
    \caption{Visualization of our AGMA on Real3D-AD. AGAM extracts geometric variation information from the neighborhood of points, with regions colored in blue indicating gentle changes in geometric structure and areas colored in red exhibiting drastic changes.}
    \label{AGMA_vis}
\end{figure}

\section{Conclusion}
In this paper, we propose a unified reconstruction framework for multi-category anomaly detection. We introduce an adaptive geometry-aware guided mask attention module, where geometric variation information is captured for robust and generalized representation of different categories. Additionally, we design a geometry-aware transformer with global position embeddings and local mask attention to learn robust reconstructed features. Experiments on benchmark datasets show that our method outperforms existing approaches, achieving state-of-the-art performance. However, further research is needed to exploit the utilization of geometric variation information and develop more robust and efficient frameworks for multi-category anomaly detection.

\section*{Acknowledgments}
This work was supported in part by the National Natural Science Foundation of China (Grant Nos. 62476171, 62476172, and 62206122), the Guangdong Basic and Applied Basic Research Foundation (Grant No. 2024A1515011367), the Guangdong Provincial Key Laboratory (Grant No. 2023B1212060076), the Tencent ``Rhinoceros Birds” - Scientific Research Foundation for Young Teachers of Shenzhen University, and the Shenzhen Institute of Artificial Intelligence and Robotics for Society.
\bibliographystyle{named}
\bibliography{ijcai25}

\begin{thebibliography}{}

\bibitem[\protect\citeauthoryear{Bengio \bgroup \em et al.\egroup }{2013}]{noise}
Yoshua Bengio, Li~Yao, Guillaume Alain, and Pascal Vincent.
\newblock Generalized denoising auto-encoders as generative models.
\newblock In {\em Proceedings of the Advances in Neural Information Processing Systems}, pages 899--907, 2013.

\bibitem[\protect\citeauthoryear{Bergmann and Sattlegger}{2023}]{Bergmann_2023_3DST}
Paul Bergmann and David Sattlegger.
\newblock Anomaly detection in 3d point clouds using deep geometric descriptors.
\newblock In {\em Proceedings of the IEEE/CVF Winter Conference on Applications of Computer Vision (WACV)}, pages 2613--2623, 2023.

\bibitem[\protect\citeauthoryear{Bergmann \bgroup \em et al.\egroup }{2022}]{Bergmann_2022}
Paul Bergmann, Xin Jin, David Sattlegger, and Carsten Steger.
\newblock The mvtec 3d-ad dataset for unsupervised 3d anomaly detection and localization.
\newblock In {\em Proceedings of the 17th International Joint Conference on Computer Vision, Imaging and Computer Graphics Theory and Applications}, pages 202--213, 2022.

\bibitem[\protect\citeauthoryear{Cao \bgroup \em et al.\egroup }{2024}]{cao2023complementarypseudomultimodalfeature}
Yunkang Cao, Xiaohao Xu, and Weiming Shen.
\newblock Complementary pseudo multimodal feature for point cloud anomaly detection.
\newblock {\em Pattern Recognition}, 156:110761, 2024.

\bibitem[\protect\citeauthoryear{Chu \bgroup \em et al.\egroup }{2023}]{shape2023chu}
Yu-Min Chu, Chieh Liu, Ting-I Hsieh, Hwann-Tzong Chen, and Tyng-Luh Liu.
\newblock Shape-guided dual-memory learning for 3{D} anomaly detection.
\newblock In {\em Proceedings of the 40th International Conference on Machine Learning}, pages 6185--6194, 2023.

\bibitem[\protect\citeauthoryear{He \bgroup \em et al.\egroup }{2024}]{Diad}
Haoyang He, Jiangning Zhang, Hongxu Chen, Xuhai Chen, Zhishan Li, Xu~Chen, Yabiao Wang, Chengjie Wang, and Lei Xie.
\newblock A diffusion-based framework for multi-class anomaly detection.
\newblock In {\em Proceedings of the AAAI Conference on Artificial Intelligence}, pages 8472--8480, 2024.

\bibitem[\protect\citeauthoryear{Horwitz and Hoshen}{2023}]{Horwitz2023BTF}
Eliahu Horwitz and Yedid Hoshen.
\newblock Back to the feature: Classical 3d features are (almost) all you need for 3d anomaly detection.
\newblock In {\em Proceedings of the IEEE/CVF Conference on Computer Vision and Pattern Recognition Workshops}, pages 2968--2977, 2023.

\bibitem[\protect\citeauthoryear{Li \bgroup \em et al.\egroup }{2023}]{Li_2023_CVPR}
Zechuan Li, Hongshan Yu, Zhengeng Yang, Tongjia Chen, and Naveed Akhtar.
\newblock Ashapeformer: Semantics-guided object-level active shape encoding for 3d object detection via transformers.
\newblock In {\em Proceedings of the IEEE/CVF Conference on Computer Vision and Pattern Recognition}, pages 1012--1021, 2023.

\bibitem[\protect\citeauthoryear{Li \bgroup \em et al.\egroup }{2024}]{IMRNet}
Wenqiao Li, Xiaohao Xu, Yao Gu, Bozhong Zheng, Shenghua Gao, and Yingna Wu.
\newblock Towards scalable 3d anomaly detection and localization: A benchmark via 3d anomaly synthesis and a self-supervised learning network.
\newblock In {\em Proceedings of the 2024 IEEE/CVF Conference on Computer Vision and Pattern Recognition}, pages 22207--22216, 2024.

\bibitem[\protect\citeauthoryear{Liang \bgroup \em et al.\egroup }{2025}]{LookLiang}
Hanzhe Liang, Guoyang Xie, Chengbin Hou, Bingshu Wang, Can Gao, and Jinbao Wang.
\newblock Look inside for more: Internal spatial modality perception for 3d anomaly detection.
\newblock In {\em Proceedings of the AAAI Conference on Artificial Intelligence}, pages 5146--5154, 2025.

\bibitem[\protect\citeauthoryear{Liu \bgroup \em et al.\egroup }{2023}]{real3dad}
Jiaqi Liu, Guoyang Xie, Ruitao Chen, Xinpeng Li, Jinbao Wang, Yong Liu, Chengjie Wang, and Feng Zheng.
\newblock Real3d-ad: A dataset of point cloud anomaly detection.
\newblock In {\em Proceedings of the Advances in Neural Information Processing Systems}, pages 30402--30415, 2023.

\bibitem[\protect\citeauthoryear{Lu \bgroup \em et al.\egroup }{2023}]{Lu2023hvq}
Ruiying Lu, YuJie Wu, Long Tian, Dongsheng Wang, Bo~Chen, Xiyang Liu, and Ruimin Hu.
\newblock Hierarchical vector quantized transformer for multi-class unsupervised anomaly detection.
\newblock In {\em Proceedings of the Advances in Neural Information Processing Systems}, pages 8487--8500, 2023.

\bibitem[\protect\citeauthoryear{Pang \bgroup \em et al.\egroup }{2022}]{Pang2022pointmae}
Yatian Pang, Wenxiao Wang, Francis E.~H. Tay, Wei Liu, Yonghong Tian, and Li~Yuan.
\newblock Masked autoencoders for point cloud self-supervised learning.
\newblock In {\em Proceedings of the 16th European Conference Computer Vision}, pages 604--621, 2022.

\bibitem[\protect\citeauthoryear{Roth \bgroup \em et al.\egroup }{2022}]{patchcore}
Karsten Roth, Latha Pemula, Joaquin Zepeda, Bernhard Sch{\"o}lkopf, Thomas Brox, and Peter Gehler.
\newblock Towards total recall in industrial anomaly detection.
\newblock In {\em Proceedings of the IEEE/CVF Conference on Computer Vision and Pattern Recognition}, pages 14318--14328, 2022.

\bibitem[\protect\citeauthoryear{Vaswani \bgroup \em et al.\egroup }{2017}]{transformer}
Ashish Vaswani, Noam Shazeer, Niki Parmar, Jakob Uszkoreit, Llion Jones, Aidan~N Gomez, \L~ukasz Kaiser, and Illia Polosukhin.
\newblock Attention is all you need.
\newblock In {\em Proceedings of the Advances in Neural Information Processing Systems}, pages 5998--6008, 2017.

\bibitem[\protect\citeauthoryear{Wang \bgroup \em et al.\egroup }{2023}]{Wang2023multimodal}
Yue Wang, Jinlong Peng, Jiangning Zhang, Ran Yi, Yabiao Wang, and Chengjie Wang.
\newblock Multimodal industrial anomaly detection via hybrid fusion.
\newblock In {\em Proceedings of the IEEE/CVF Conference on Computer Vision and Pattern Recognition}, pages 8032--8041, 2023.

\bibitem[\protect\citeauthoryear{Wu \bgroup \em et al.\egroup }{2015}]{modelnet40}
Zhirong Wu, Shuran Song, Aditya Khosla, Fisher Yu, Linguang Zhang, Xiaoou Tang, and Jianxiong Xiao.
\newblock 3d shapenets: A deep representation for volumetric shapes.
\newblock In {\em Proceedings of the IEEE Conference on Computer Vision and Pattern Recognition}, pages 1912--1920, 2015.

\bibitem[\protect\citeauthoryear{Ye \bgroup \em et al.\egroup }{2024}]{PO3AD}
Jianan Ye, Weiguang Zhao, Xi~Yang, Guangliang Cheng, and Kaizhu Huang.
\newblock Po3ad: Predicting point offsets toward better 3d point cloud anomaly detection.
\newblock {\em arXiv: 2412.12617}, pages 1--13, 2024.

\bibitem[\protect\citeauthoryear{You \bgroup \em et al.\egroup }{2022}]{uniad}
Zhiyuan You, Lei Cui, Yujun Shen, Kai Yang, Xin Lu, Yu~Zheng, and Xinyi Le.
\newblock A unified model for multi-class anomaly detection.
\newblock In {\em Proceedings of the Advances in Neural Information Processing Systems}, pages 4571--4584, 2022.

\bibitem[\protect\citeauthoryear{Zavrtanik \bgroup \em et al.\egroup }{2021}]{DRAEM}
Vitjan Zavrtanik, Matej Kristan, and Danijel Skočaj.
\newblock DrÆm – a discriminatively trained reconstruction embedding for surface anomaly detection.
\newblock In {\em Proceedings of the 2021 IEEE/CVF International Conference on Computer Vision}, pages 8310--8319, 2021.

\bibitem[\protect\citeauthoryear{Zhang \bgroup \em et al.\egroup }{2023}]{DeSTSeg2023}
Xuan Zhang, Shiyu Li, Xi~Li, Ping Huang, Jiulong Shan, and Ting Chen.
\newblock Destseg: Segmentation guided denoising student-teacher for anomaly detection.
\newblock In {\em Proceedings of the 2023 IEEE/CVF Conference on Computer Vision and Pattern Recognition}, pages 3914--3923, 2023.

\bibitem[\protect\citeauthoryear{Zhou \bgroup \em et al.\egroup }{2024a}]{pointad2024}
Qihang Zhou, Jiangtao Yan, Shibo He, Wenchao Meng, and Jiming Chen.
\newblock Pointad: Comprehending 3d anomalies from points and pixels for zero-shot 3d anomaly detection.
\newblock In {\em Proceedings of the Advances in Neural Information Processing Systems}, pages 84866--84896, 2024.

\bibitem[\protect\citeauthoryear{Zhou \bgroup \em et al.\egroup }{2024b}]{Zhou2024R3D}
Zheyuan Zhou, Le~Wang, Naiyu Fang, Zili Wang, Lemiao Qiu, and Shuyou Zhang.
\newblock R3d-ad: Reconstruction via diffusion for 3d anomaly detection.
\newblock In {\em Proceedings of the 18th European Conference Computer Vision}, pages 91--107, 2024.

\bibitem[\protect\citeauthoryear{Zhu \bgroup \em et al.\egroup }{2024}]{zhu2024towards}
Hongze Zhu, Guoyang Xie, Chengbin Hou, Tao Dai, Can Gao, Jinbao Wang, and Linlin Shen.
\newblock Towards high-resolution 3d anomaly detection via group-level feature contrastive learning.
\newblock In {\em Proceedings of the 32nd ACM International Conference on Multimedia}, pages 4680--4689, 2024.

\end{thebibliography}

\newpage
\appendix

\end{document}


\maketitle

\section{Additional Experiments}
\subsection{Experiment Results}
\textbf{Results Anomaly-ShapeNet.} The pixel-level AUROC results of MC3D-AD on the Anomaly-ShapeNet dataset~\cite{IMRNet} are presented in Table~\ref{Anomaly-ShapeNet-p}. It can be observed that MC3D-AD achieves an 8.0\% improvement in pixel-level AUROC compared to the second single-category method. This further confirms the effectiveness of MC3D-AD in multi-category anomaly localization.

\begin{table*}[!b]
  \centering
\resizebox{\textwidth}{!}{

    \begin{tabular}{l|cccccccccccccc}
    \toprule
    \multicolumn{15}{c}{\textbf{P-AUROC($\uparrow$)}} \\
    \midrule
    \textbf{Method} & \textbf{cap0} & \textbf{cap3} & \textbf{helmet3} & \textbf{cup0} & \textbf{bowl4} & \textbf{vase3} & \textbf{headset1} & \textbf{eraser0} & \textbf{vase8} & \textbf{cap4} & \textbf{vase2} & \textbf{vase4} & \textbf{helmet0} & \textbf{bucket1} \\
    \midrule
    \textbf{BTF(Raw)}            & 0.524                                 & 0.687                                 & 0.700                                 & 0.632                                 & 0.563                                 & 0.602                                 & 0.475                                 & 0.637                                 & 0.550                                 & 0.469                                 & 0.403                                 & 0.613                                 & 0.504                                 & 0.686                                 \\
\textbf{BTF(FPFH)}           & {\color[HTML]{0070C0} \textbf{0.730}} & 0.658                                 & {\color[HTML]{0070C0} \textbf{0.724}} & {\color[HTML]{FF0000} \textbf{0.790}} & 0.679                                 & {\color[HTML]{0070C0} \textbf{0.699}} & 0.591                                 & 0.719                                 & 0.662                                 & 0.524                                 & 0.646                                 & 0.710                                 & 0.575                                 & 0.633                                 \\
\textbf{M3DM}                & 0.531                                 & 0.605                                 & 0.655                                 & 0.715                                 & 0.624                                 & 0.658                                 & 0.585                                 & 0.710                                 & 0.551                                 & 0.718                                 & 0.737                                 & 0.655                                 & 0.599                                 & 0.699                                 \\
\textbf{Patchcore(FPFH)}     & 0.472                                 & 0.653                                 & {\color[HTML]{FF0000} \textbf{0.737}} & 0.655                                 & {\color[HTML]{0070C0} \textbf{0.720}} & 0.430                                 & 0.464                                 & {\color[HTML]{0070C0} \textbf{0.810}} & 0.575                                 & 0.595                                 & 0.721                                 & 0.505                                 & 0.548                                 & 0.571                                 \\
\textbf{Patchcore(PointMAE)} & 0.544                                 & 0.488                                 & 0.615                                 & 0.510                                 & 0.501                                 & 0.465                                 & 0.423                                 & 0.378                                 & 0.364                                 & 0.725                                 & {\color[HTML]{0070C0} \textbf{0.742}} & 0.523                                 & 0.580                                 & 0.574                                 \\
\textbf{CPMF}                & 0.601                                 & 0.551                                 & 0.520                                 & 0.497                                 & 0.683                                 & 0.582                                 & 0.458                                 & 0.689                                 & 0.529                                 & 0.553                                 & 0.582                                 & 0.514                                 & 0.555                                 & 0.601                                 \\
\textbf{Reg3D-AD}               & 0.632                                 & {\color[HTML]{0070C0} \textbf{0.718}} & 0.620                                 & 0.685                                 & {\color[HTML]{FF0000} \textbf{0.800}} & 0.511                                 & {\color[HTML]{FF0000} \textbf{0.626}} & 0.755                                 & {\color[HTML]{0070C0} \textbf{0.811}} & {\color[HTML]{0070C0} \textbf{0.815}} & 0.405                                 & {\color[HTML]{0070C0} \textbf{0.755}} & {\color[HTML]{0070C0} \textbf{0.600}} & 0.752                                 \\
\textbf{IMRNet}              & 0.715                                 & 0.706                                 & 0.663                                 & 0.643                                 & 0.576                                 & 0.401                                 & 0.476                                 & 0.548                                 & 0.635                                 & 0.753                                 & 0.614                                 & 0.524                                 & 0.598                                 & {\color[HTML]{0070C0} \textbf{0.774}} \\
\textbf{Ours}                & {\color[HTML]{FF0000} \textbf{0.854}} & {\color[HTML]{FF0000} \textbf{0.903}} & 0.585                                 & {\color[HTML]{0070C0} \textbf{0.763}} & 0.670                                 & {\color[HTML]{FF0000} \textbf{0.800}} & {\color[HTML]{0070C0} \textbf{0.592}} & {\color[HTML]{FF0000} \textbf{0.820}} & {\color[HTML]{FF0000} \textbf{0.874}} & {\color[HTML]{FF0000} \textbf{0.858}} & {\color[HTML]{FF0000} \textbf{0.781}} & {\color[HTML]{FF0000} \textbf{0.772}} & {\color[HTML]{FF0000} \textbf{0.749}} & {\color[HTML]{FF0000} \textbf{0.868}} \\
    \midrule
    \multicolumn{1}{c}{} &       &       &       &       &       &       &       &       &       &       &       &       &       &  \\
    \midrule
    \textbf{Method} & \textbf{bottle3} & \textbf{vase0} & \textbf{bottle0} & \textbf{tap1} & \textbf{bowl0} & \textbf{bucket0} & \textbf{vase5} & \textbf{vase1} & \textbf{vase9} & \textbf{ashtray0} & \textbf{bottle1} & \textbf{tap0} & \textbf{phone} & \textbf{cup1} \\
    \midrule
    \textbf{BTF(Raw)}            & {\color[HTML]{0070C0} \textbf{0.720}} & 0.618                                 & 0.551                                 & 0.564                                 & 0.524                                 & 0.617                                 & 0.585                                 & 0.549                                 & 0.564                                 & 0.512                                 & 0.491                                 & 0.527                                 & 0.583                                 & 0.561                                 \\
\textbf{BTF(FPFH)}           & 0.622                                 & 0.642                                 & 0.641                                 & 0.596                                 & 0.710                                 & 0.401                                 & 0.429                                 & {\color[HTML]{0070C0} \textbf{0.619}} & 0.568                                 & 0.624                                 & 0.549                                 & 0.568                                 & 0.675                                 & 0.619                                 \\
\textbf{M3DM}                & 0.532                                 & 0.608                                 & 0.663                                 & 0.712                                 & 0.658                                 & {\color[HTML]{0070C0} \textbf{0.698}} & 0.642                                 & 0.602                                 & 0.663                                 & 0.577                                 & 0.637                                 & 0.654                                 & 0.358                                 & 0.556                                 \\
\textbf{Patchcore(FPFH)}     & 0.512                                 & 0.655                                 & 0.654                                 & {\color[HTML]{FF0000} \textbf{0.768}} & 0.524                                 & 0.459                                 & 0.447                                 & 0.453                                 & 0.663                                 & 0.597                                 & 0.687                                 & {\color[HTML]{0070C0} \textbf{0.733}} & 0.488                                 & 0.596                                 \\
\textbf{Patchcore(PointMAE)} & 0.653                                 & {\color[HTML]{0070C0} \textbf{0.677}} & 0.553                                 & 0.541                                 & 0.527                                 & 0.586                                 & 0.572                                 & 0.551                                 & 0.423                                 & 0.495                                 & 0.606                                 & {\color[HTML]{FF0000} \textbf{0.858}} & {\color[HTML]{0070C0} \textbf{0.886}} & {\color[HTML]{FF0000} \textbf{0.856}} \\
\textbf{CPMF}                & 0.435                                 & 0.458                                 & 0.521                                 & 0.657                                 & 0.745                                 & 0.486                                 & {\color[HTML]{0070C0} \textbf{0.651}} & 0.486                                 & 0.545                                 & 0.615                                 & 0.571                                 & 0.458                                 & 0.545                                 & 0.509                                 \\
\textbf{Reg3D-AD}               & 0.525                                 & 0.548                                 & {\color[HTML]{0070C0} \textbf{0.886}} & {\color[HTML]{0070C0} \textbf{0.741}} & {\color[HTML]{0070C0} \textbf{0.775}} & 0.619                                 & 0.624                                 & 0.602                                 & {\color[HTML]{0070C0} \textbf{0.694}} & {\color[HTML]{0070C0} \textbf{0.698}} & 0.696                                 & 0.589                                 & 0.599                                 & {\color[HTML]{0070C0} \textbf{0.698}} \\
\textbf{IMRNet}              & 0.641                                 & 0.535                                 & 0.556                                 & 0.699                                 & {\color[HTML]{FF0000} \textbf{0.781}} & 0.585                                 & {\color[HTML]{FF0000} \textbf{0.682}} & {\color[HTML]{FF0000} \textbf{0.685}} & 0.691                                 & 0.671                                 & {\color[HTML]{0070C0} \textbf{0.702}} & 0.681                                 & 0.742                                 & 0.688                                 \\
\textbf{Ours}                & {\color[HTML]{FF0000} \textbf{0.902}} & {\color[HTML]{FF0000} \textbf{0.897}} & {\color[HTML]{FF0000} \textbf{0.902}} & 0.584                                 & {\color[HTML]{0070C0} \textbf{0.775}} & {\color[HTML]{FF0000} \textbf{0.902}} & 0.588                                 & 0.608                                 & {\color[HTML]{FF0000} \textbf{0.762}} & {\color[HTML]{FF0000} \textbf{0.807}} & {\color[HTML]{FF0000} \textbf{0.867}} & 0.502                                 & {\color[HTML]{FF0000} \textbf{0.891}} & 0.694 \\
    \midrule
    \multicolumn{1}{c}{} &       &       &       &       &       &       &       &       &       &       &       &       &       &  \\
    \midrule
    \textbf{Method} & \textbf{vase7} & \textbf{helmet2} & \textbf{cap5} & \textbf{shelf0} & \textbf{bowl5} & \textbf{bowl3} & \textbf{helmet1} & \textbf{bowl1} & \textbf{headset0} & \textbf{bag0} & \textbf{bowl2} & \textbf{jar} & \multicolumn{2}{|c}{\textbf{Mean}} \\
    \midrule
    \textbf{BTF(FPFH)}           & 0.540                                 & 0.643                                 & 0.586                                 & 0.619                                 & {\color[HTML]{0070C0} \textbf{0.699}} & 0.590                                 & {\color[HTML]{FF0000} \textbf{0.749}} & {\color[HTML]{FF0000} \textbf{0.768}} & 0.620                                 & {\color[HTML]{0070C0} \textbf{0.746}} & 0.518                                 & 0.427                                 & \multicolumn{2}{|c}{0.628}                                 \\
\textbf{M3DM}                & 0.517                                 & 0.623                                 & 0.655                                 & 0.554                                 & 0.489                                 & 0.657                                 & 0.427                                 & 0.663                                 & 0.581                                 & 0.637                                 & {\color[HTML]{FF0000} \textbf{0.694}} & 0.541                                 & \multicolumn{2}{|c}{0.616}                                 \\
\textbf{Patchcore(FPFH)}     & {\color[HTML]{0070C0} \textbf{0.693}} & 0.455                                 & {\color[HTML]{0070C0} \textbf{0.795}} & 0.613                                 & 0.358                                 & 0.327                                 & 0.489                                 & 0.531                                 & 0.583                                 & 0.574                                 & 0.625                                 & 0.478                                 & \multicolumn{2}{|c}{0.580}                                 \\
\textbf{Patchcore(PointMAE)} & 0.651                                 & 0.651                                 & 0.545                                 & 0.543                                 & 0.562                                 & 0.581                                 & 0.562                                 & 0.524                                 & 0.575                                 & 0.674                                 & 0.515                                 & 0.487                                 & \multicolumn{2}{|c}{0.577}                                 \\
\textbf{CPMF}                & 0.504                                 & 0.515                                 & 0.551                                 & {\color[HTML]{FF0000} \textbf{0.783}} & 0.684                                 & 0.641                                 & 0.542                                 & 0.488                                 & {\color[HTML]{0070C0} \textbf{0.699}} & 0.655                                 & 0.635                                 & 0.611                                 & \multicolumn{2}{|c}{0.573}                                 \\
\textbf{Reg3D-AD}               & {\color[HTML]{FF0000} \textbf{0.881}} & {\color[HTML]{FF0000} \textbf{0.825}} & 0.467                                 & {\color[HTML]{0070C0} \textbf{0.688}} & 0.691                                 & 0.654                                 & {\color[HTML]{0070C0} \textbf{0.624}} & 0.615                                 & 0.580                                 & 0.715                                 & 0.593                                 & 0.599                                 & \multicolumn{2}{|c}{{\color[HTML]{0070C0} \textbf{0.668}}} \\
\textbf{IMRNet}              & 0.593                                 & 0.644                                 & 0.742                                 & 0.605                                 & {\color[HTML]{FF0000} \textbf{0.715}} & 0.599                                 & 0.604                                 & {\color[HTML]{0070C0} \textbf{0.705}} & {\color[HTML]{FF0000} \textbf{0.705}} & 0.668                                 & {\color[HTML]{0070C0} \textbf{0.684}} & {\color[HTML]{0070C0} \textbf{0.765}} & \multicolumn{2}{|c}{0.650}                                 \\
\textbf{Ours}                & 0.576                                 & {\color[HTML]{0070C0} \textbf{0.818}} & {\color[HTML]{FF0000} \textbf{0.882}} & 0.625                                & 0.562                                 & {\color[HTML]{FF0000} \textbf{0.779}} & 0.591                                 & 0.562                                 & 0.666                                 & {\color[HTML]{FF0000} \textbf{0.857}} & 0.597                                 & {\color[HTML]{FF0000} \textbf{0.847}} & \multicolumn{2}{|c}{{\color[HTML]{FF0000} \textbf{0.748}}} \\
    \bottomrule
    \end{tabular}
}
  \caption{The pixel-level AUROC experimental results for anomaly location across 40 categories of Anomaly-ShapeNet. The best and the second-best results are highlighted in \textcolor[rgb]{1, 0, 0}{\textbf{red}}  and \textcolor[rgb]{0, .439, .753}{\textbf{blue}}, respectively. The results of the baselines are excerpted from their papers.}
  \label{Anomaly-ShapeNet-p}
\end{table*}

\subsection{More Ablation Results}
\begin{table}[h!]
\vspace{-12pt}
\setlength{\abovecaptionskip}{5pt}
\setlength{\belowcaptionskip}{5pt}
\resizebox{\columnwidth}{!}{
\begin{tabular}{@{}cccccc@{}}
\multicolumn{6}{c}{(a)}                                                                                         \\ \toprule
$\text{LGE}_{\textbf{w/o\ AGMA}}$          & $\text{LGE}_{\textbf{AGMA}}$   & $\text{GQD}_{\textbf{w/o\ AGMA}}$                          & $\text{GQD}_{\textbf{AGMA}}$          & O-AUROC        & P-AUROC        \\ \midrule
\checkmark   &            &                              &                    & 0.658          & 0.650          \\
\checkmark   & \checkmark &                              &                    & 0.752          & 0.709          \\
\checkmark   &            & \checkmark                   &                    & 0.741          & 0.733          \\
\checkmark   & \checkmark & \checkmark                   &                    & 0.756          & 0.755          \\
\checkmark   & \checkmark & \checkmark                   & \checkmark         & \textbf{0.782} & \textbf{0.768} \\ \bottomrule
\multicolumn{6}{c}{(b)}                                                                                         \\ \toprule
Sampling     & O-AUROC    & \multicolumn{1}{c|}{P-AUROC} & Perturbation       & O-AUROC        & P-AUROC        \\ \midrule
FPS(1024)    & 0.729      & \multicolumn{1}{c|}{0.734}   & 0(no perturbaiton) &     0.753      &   0.738         \\
FPS(2048)    &      0.747      & \multicolumn{1}{c|}{0.757}        & 5                  &      0.758       &       0.735      \\
FPS(4096)    & \textbf{0.782}      & \multicolumn{1}{c|}{\textbf{0.768}}   & 10                 &       0.771     &    0.756        \\
Random(1024) &    0.719   & \multicolumn{1}{c|}{0.724}        & 15                 &       0.774    &   0.761         \\
Random(2048) &    0.738  & \multicolumn{1}{c|}{0.746}        & 20                 &    \textbf{0.782}      &  \textbf{0.768}              \\
Random(4096) & 0.775      & \multicolumn{1}{c|}{0.756}   & 25                 &   0.766      &     0.752      \\ \bottomrule
\end{tabular}
}
\caption{More ablation results on Real3D-AD. (a) Ablation on key modules, where  $\textbf{w/o AGMA}$ means model without the AGMA module. (b) Ablation on Farthest Point Sampling (FPS) and perturbation.}
\label{moreablation}
\end{table}

\begin{figure*}[h!]
    \centering
    \includegraphics[width=0.9\linewidth]{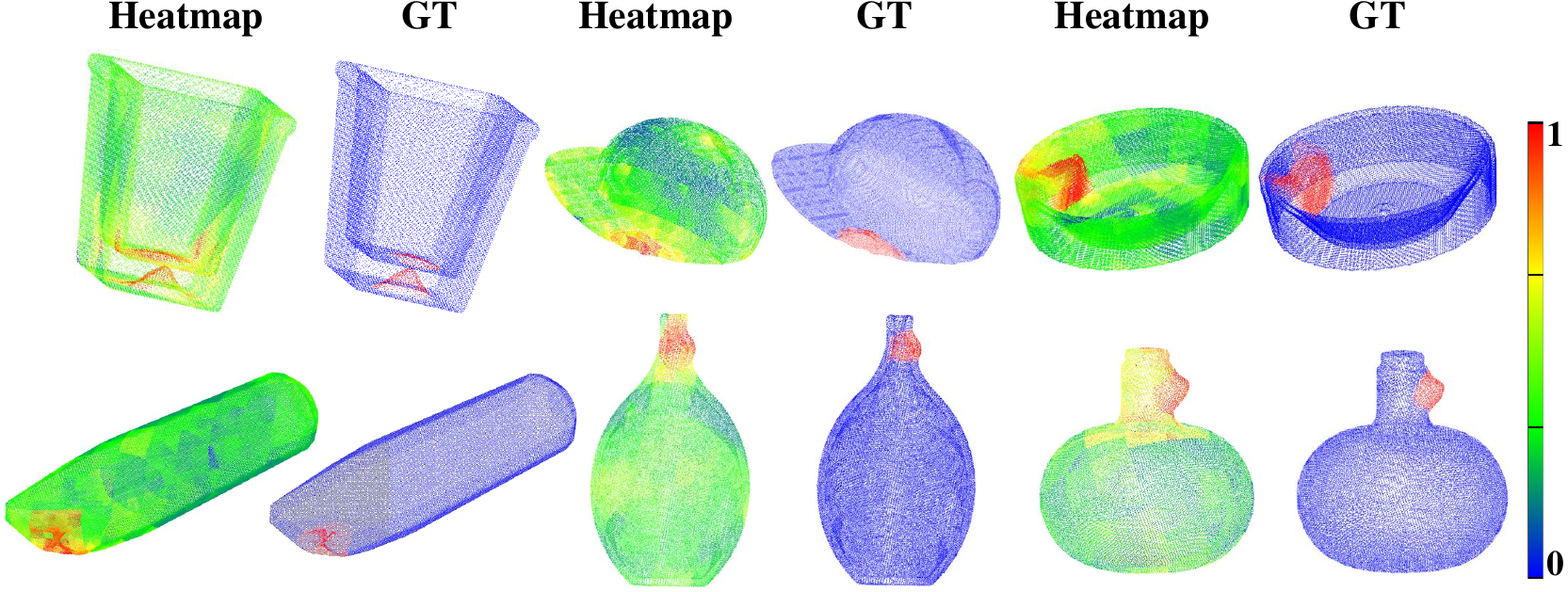}
    \caption{Point heatmap comparison of our MC3D-AD with the Ground Truth (GT) on Anomaly-ShapeNet.}
    \label{heatmap_vis_Anomaly-ShapeNet}
\end{figure*}

To further analyze the effectiveness of the proposed modules, we conducted additional ablation experiments. The results are presented in Table~\ref{moreablation}. As shown in Table~\ref{moreablation}(a), AGMA is critical for capturing the geometric structure of point clouds, leading to a substantial improvement in reconstruction quality and an improvement of 9.4\% in O-AUROC for anomaly detection. Meanwhile, LQD provides additional supervision during reconstruction, enhancing anomaly localization and achieving a gain of 8.3\% in P-AUROC. Notably, AGMA can be seamlessly integrated with both LGE and GQD, resulting in consistent performance improvements across all evaluation metrics. As shown in Table 2(b), FPS sampling better preserves the geometric structure of point clouds, and increasing the number of sampled points enables the model to learn more comprehensive geometric representations. Moreover, introducing an appropriate level of noise facilitates the learning of normal representations, thereby enhancing the model’s robustness.

\subsection{Visualization}
Figure~\ref{heatmap_vis_Anomaly-ShapeNet} shows the visualized results of our method on Anomaly-ShapeNet. It is clear that MC3D-AD can accurately detect and locate small anomalies within the point cloud from different categories.

\subsection{Generalization of Our Method for Classification}
To examine the generalization of our method, extra classification experiments are conducted on Real3D-AD~\cite{real3dad}, and the results are shown in Table~\ref{real3D_cls}. It can be observed in Table~\ref{real3D_cls} that, without losing anomaly detection performance, the classification capability remains satisfactory even when each class contains only four training samples. 
\begin{table}[!h]
\footnotesize
\centering
\begin{tabular}{@{}lccc@{}}
\toprule
\textbf{Category} & \textbf{Accuracy} & \textbf{O-AUROC} & \textbf{P-AUROC} \\ \midrule
car               & 1.000             & 0.700            & 0.816           \\
shell             & 0.881             & 0.820            & 0.792           \\
fish              & 0.843             & 0.863            & 0.943           \\
chicken           & 0.623             & 0.722            & 0.607           \\
diamond           & 0.910             & 0.756            & 0.831           \\
candybar          & 0.765             & 0.839            & 0.969           \\
starfish          & 0.897             & 0.758            & 0.661           \\
toffees           & 0.442             & 0.794            & 0.896           \\
duck              & 0.918             & 0.724            & 0.847           \\
seahorse          & 0.983             & 0.687            & 0.667           \\
airplane          & 0.932             & 0.864            & 0.621           \\
gemstone          & 1.000             & 0.475            & 0.385           \\ \midrule
mean              & 0.849             & 0.750            & 0.753           \\ \bottomrule
\end{tabular}
\caption{Object classification and anomaly detection performance of our MC3D-AD on Real3D-AD.}
\label{real3D_cls}
\end{table}

\begin{table}[!h]
\footnotesize
\centering
\begin{tabular}{@{}lccc@{}}
\toprule
\textbf{Category} & \textbf{Accuracy} & \textbf{O-AUROC} & \textbf{PAUROC} \\ \midrule
bowl4             & 0.879             & 0.641            & 0.555           \\
cup0              & 0.931             & 0.924            & 0.694           \\
bucket0           & 0.528             & 0.911            & 0.639           \\
bottle0           & 1.000             & 0.800            & 0.775           \\
tap1              & 1.000             & 0.944            & 0.522           \\
headset1          & 1.000             & 0.838            & 0.571           \\
vase3             & 0.973             & 0.824            & 0.699           \\
helmet3           & 0.838             & 0.976            & 0.624           \\
shelf0            & 0.895             & 0.783            & 0.592           \\
cap0              & 0.970             & 0.737            & 0.763           \\ \midrule
mean              & 0.901             & 0.838            & 0.643           \\ \bottomrule
\end{tabular}
\caption{Object classification and anomaly detection performance of our MC3D-AD on Anomaly-ShapeNet}
\label{ShapeNet_cls}
\end{table}

The Anomaly-ShapeNet dataset contains 40 categories, with only 4 training samples per class, making the classification task significantly more challenging. Therefore, experiments are conducted on a subset of the dataset with 10 categories and the experimental results are shown in Table~\ref{ShapeNet_cls}. It is evident that our MC3D-AD achieved an accuracy of 0.901, which further demonstrates the effectiveness of our MC3D-AD in dealing with multi-task. 

\subsection{Extensibility of the Proposed AGMA}
To evaluate the extensibility of AGMA, it was integrated into PointMAE~\cite{Pang2022pointmae} to perform point cloud classification and segmentation tasks on ModelNet40~\cite{modelnet40} and ShapeNet-part~\cite{shapenetpart}. The methods selected for comparison include PointNet~\cite{pointnet}, PointNet++~\cite{pointnet++}, DGCNN~\cite{DGCNN}, and PointMAE~\cite{Pang2022pointmae}. The experimental results are shown in Table~\ref{cls_modelnet} and Table~\ref{partseg}, respectively.
\begin{table}[h!]
\centering
\begin{tabular}{lc}
\hline
\textbf{Method} & \textbf{Accuracy} \\ \hline
PointNet        & 0.892             \\
PointNet++      & 0.907             \\
DGCNN           & 0.929             \\
PointMAE        & 0.931             \\
PointMAE$_{\text{AGMA}}$ & 0.934             \\ \hline
\end{tabular}
\caption{Point cloud classification performance on ModelNet40.}
\label{cls_modelnet}
\end{table}

\begin{table}[h]
\centering
\begin{tabular}{lc}
\hline
\textbf{Method} & \textbf{IoU} \\ \hline
PointNet        & 0.837        \\
PointNet++      & 0.851        \\
DGCNN           & 0.852        \\
PointMAE        & 0.860        \\
PointMAE$_{\text{AGMA}}$ & 0.861        \\ \hline
\end{tabular}
\caption{Point cloud segmentation performance on ShapeNet-part.}
\label{partseg}
\end{table}
The experimental results show that the performance improvement is highly related to the number of point cloud groups. For ShapeNet\_part, where the group parameter is set to 128, the improvement of instance average Intersection
over Union (IoU) is modest, while for ModelNet40, with a group size of 512, the accuracy is clearly improved. This is because sparse group centers contain insufficient geometric information, making it challenging for AGAM to capture the spatial structure of the point cloud.

\subsection{Inference Speed}
\begin{figure}[!h]
    \centering
    \includegraphics[width=.7\linewidth]{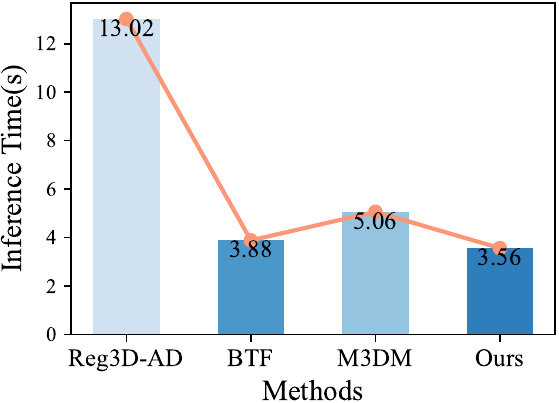}
    \caption{Average inference time ($\downarrow$) per object on Real3D-AD}
    \label{inference_time}
\end{figure}

In real-world scenarios, inference speed is very important for model deployment, so relevant experiments are conducted and the experimental results are shown in Fig~\ref{inference_time}.

It is clear that our method outperforms BTF~\cite{Horwitz2023BTF}, M3DM~\cite{Wang2023multimodal}, and Reg3D-AD~\cite{real3dad} in terms of inference speed.Our method with one RTX 3090 achieved an average inference time of 3.560 seconds for each object in high-precision Real3D-AD, which is better than open-source methods such as BTF (3.882),
M3DM (5.061), and Reg3D-AD (13.022). Although M3DM and Reg3D-AD achieve good anomaly detection performance, their inference speeds are considerably slow. BTF demonstrates good efficiency, but its anomaly detection performance needs more improvement. The proposed MC3D-AD effectively balances anomaly detection performance with inference speed, highlighting its promising potential for real-world industrial applications.

\bibliographystyle{named}
\bibliography{ijcai25}